# VID-WIN: Fast Video Event Matching with Query-Aware Windowing at the Edge for the Internet of Multimedia Things

Piyush Yadav ⓘ, Dhaval Salwala ⓘ, and Edward Curry ⓘ

*Abstract*— **Efficient video processing is a critical component in many IoMT applications to detect events of interest. Presently, many window optimization techniques have been proposed in event processing with an underlying assumption that the incoming stream has a structured data model. Videos are highly complex due to the lack of any underlying structured data model. Video stream sources such as CCTV cameras and smartphones are resource-constrained edge nodes. At the same time, video content extraction is expensive and requires computationally intensive Deep Neural Network (DNN) models that are primarily deployed at high-end (or cloud) nodes. This paper presents VID-WIN, an adaptive 2-stage allied windowing approach to accelerate video event analytics in an edge-cloud paradigm. VID-WIN runs parallelly across edge and cloud nodes and performs the query and resource-aware optimization for state-based complex event matching. VID-WIN exploits the video content and DNN input knobs to accelerate the video inference process across nodes. The paper proposes a novel content-driven *micro-batch resizing*, query-aware *caching* and micro-batch based *utility filtering* strategy of video frames under resource-constrained edge nodes to improve the overall system throughput, latency, and network usage. Extensive evaluations are performed over five real-world datasets. The experimental results show that VID-WIN video event matching achieves ~2.3X higher throughput with minimal latency and ~99% bandwidth reduction compared to other baselines while maintaining query-level accuracy and resource bounds.**

## I. INTRODUCTION

With the advancement in the Internet of Multimedia Things (IoMT) [1], there is a significant increase in the proliferation of visual sensors (e.g. CCTV cameras, smartphones). These visual sensors are now pervasive and generate a massive amount of video data streams. Video event processing is the key requirement in many IoMT applications like traffic analytics [2, 3], wildlife monitoring [4], object identification [2] and require responses in real-time. Complex Event Processing is an event-driven paradigm that detects patterns in real-time by performing event matching over data streams [5-8]. CEP decompose complex prediction tasks into a directed acyclic graph (DAG) of operators [9]. Depending on the application scenario, the CEP operators can be deployed on a single machine, shared memory [10], edge or cloud node [11, 12]. This modularity of operator placement helps in realizing the *distributed intelligence* in CEP systems. Recently, specialized video CEP systems have been proposed to detect event patterns over video streams [8]. These CEP system pass video streams to machine learning operators such as Deep Neural Network (DNN) object detection models (e.g. YOLO [13]) to enable complex video event pattern detection. CEP and other data stream processing systems use *window* operators to capture monotonically growing infinite data streams. In CEP, the windows operator discretizes continuous data streams as *state* and apply event rules over them to detect pattern**s**. The windows continuously accept new input and discard old data as they become irrelevant for query analysis.

Edge-based deployment and techniques are gaining vast importance in pervasive and distributed event processing. IoMT applications like video event processing require short responses for real-time pattern matching. Performing video event analytics at edge devices face challenges in terms of Quality of Service (QoS) metrics like latency, throughput, bandwidth, and accuracy due to their computing limitations. This paper presents a new CEP windowing technique for faster video event matching in the edge-cloud computing continuum. The work proposes VID-WIN, a query and resource-aware windows for video streams. VID-WIN optimizes the CEP matching performance under resource and application-level constraints for a given video pattern query.

### A. Contribution

To the best of our knowledge, this is the first work that proposes a window-based query-aware optimization over video streams for the edge-cloud paradigm. Overall, the following contributions are presented and evaluated in this paper:

1. A 2-stage *allied windowing* concept VID-WIN is proposed which runs concurrently over edge and cloud nodes to perform state-based video event matching. The 2-stage enables pre-processing and post-processing window-based optimizations for video.

2. *Content-driven* and *Query-aware* optimization on video data via *input transformation techniques* is proposed. The VID-WIN controller considers the nature of *CEP queries* such as pattern *accuracy* and *partial matches* and exploits low-level video content to accelerate DNN model inference.

3. A dynamic *micro-batching* and *resizing* approach for video frames to improve throughput and minimize end-to-end latency with reduced bandwidth usage.

4. The work proposes a novel *query-aware* partial match

This article has been published in the IEEE Internet of Things Journal. https://doi.org/10.1109/JIOT.2021.3075336. The work is funded from Science Foundation of Ireland grant SFI/12/RC/2289_P2.
P. Yadav, D. Salwala and E.Curry are with Insight SFI Research Centre for Data Analytics, Data Science Institute, National University of Ireland Galway, Ireland (e-mail: {piyush.yadav; dhaval.salwala; edward.curry}7@nuigalway.ie).

*caching* and a *micro-batch utility score* which helps in performing *state-based and resource-aware filtering* of video data to minimize network usage under limited system resources (CPU, memory) while maintaining the quality of results.

5. Large-scale experiments are performed to validate the efficacy of the VID-WIN approach over five real-world video datasets. The performance is measured across other vision analytics baselines like CloudSeg [14], Reducto [15] and CEP system such as Esper [16]. The micro-batch resizing approach achieves 30-40% higher throughput with minimal latency. The query-based cache and micro-batch utility achieve ~78.6% filtering for 50% edge resource usage with 80-99% bandwidth savings.

The rest of the paper is organized as follows: Section II presents the motivation, challenges, and proposed approach. Section III covers the background. Section IV and V discuss the tunable parameters for DNN models and conceptualize the VID-WIN windows operator. Section VI explains the VID-WIN adaptation strategy and architecture. Section VII discusses the VID-WIN query aware filtering under resource constraints scenario. Section VIII details the implementation, datasets, and experimental evaluation and limitation of this work. Section IX covers related work, and the paper concludes in Section X.

## II. MOTIVATION AND PROPOSED APPROACH

This section delves deeper into motivation and challenges and throws light on the problem statement and proposed approach.

### A. Motivational Scenario

Fig.1 shows a traditional approach where a CEP operator graph is deployed over edge and cloud nodes to process queries Query 1 and Query 2. As per Query 1, if the average temperature (*avgtemp*) is greater than 50°C within a time window of 5 sec, then the CEP system must notify the user of the rising temperature alert. This is known as *state-based* event matching where patterns are assessed over a window duration like 5 sec. In Query 2, a user is interested in counting the number of 'Car' objects over a video stream within a window of 5 seconds with Top-2 accuracy. Processing IoMT data streams like videos as compared to scalar IoT streams (like temperature) in CEP leads to multiple challenges, as discussed in the remainder of this section.

### B. Current limitation of Windows for Video Streams

As shown in Fig. 1, for Query 1 matching, the temperature stream is directly passed over windows as they have a structured representation (e.g., key-value like temp=35°C) and do not require any additional pre-processing. Performing aggregation techniques like 'Average' over windows are well-understood problems in stream processing. Aggregations and optimization techniques like sharing [17], slicing (panes [18] and pairs [19]), and SWAG [20] over windows are performed with the assumption that the data it is receiving has a structured format like key-value pairs (e.g. price, temperature values). Different *content-driven* data mining techniques [21, 22] over windows are proposed where the windows continuously analyze incoming data content to optimize the performance. For example, Bifet et al. proposed ADWIN [21], a content-based adaptive windowing method to learn new scalar data sequences. These windows continuously perform the statistical analysis over the incoming data content and change their length when the data distribution changes.

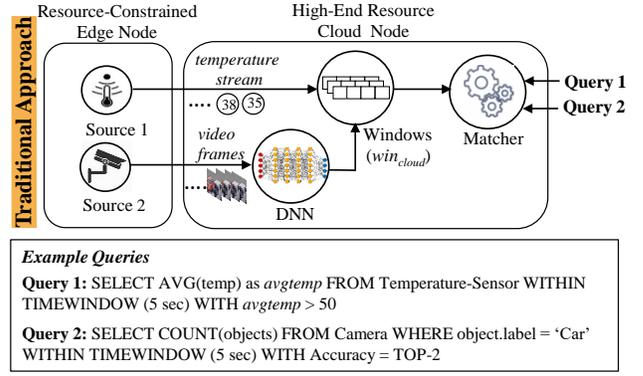

Fig. 1. Motivating Scenario: A CEP operator graph placed across different nodes for event detection of temperature and video streams.

Querying video content (such as Query 2) requires an expensive content extraction method like DNN. As shown in Fig. 1, the traditional approach is to process the video streams using DNN models and then pass the model output data over windows. However, such pre-processing misses the opportunity to exploit video and DNN model properties, which can improve the overall system performance. Currently, there are no content-driven adaptive windowing techniques proposed in the context of video data. This work focused on adaptive strategies over windows that can infer and exploit video content and DNN properties for fast event analytics in an edge-cloud setting.

### C. Challenges of Processing Video Event Queries at Edge

Many video streams are generated over IoMT edge devices which are resource-constrained in terms of memory and CPU. Deploying video analytics applications like object classification at the edge is exceptionally challenging due to the following reasons:

#### 1) Resource Level Edge Constraints

*Limited CPU and Memory:* DNN models are costly and have very low performance in resource-constrained edge devices. For example, the VGG-19 object classification model (on image resolution of 224*224) performs at the rate of 10 frames per second (fps) and 5 fps on Nvidia Jetson Nano and Raspberry Pi 3 (with Intel Neural Compute Stick 2), respectively [23]. The above are GPU powered edge devices and offer far from the usual real-time performance of ~30 fps at which the video data may be streamed. Thus, for video analytics queries (like Query 2) DNN models are primarily deployed in the high-end cloud (Fig.1) nodes as they incur high cost and are resource-intensive.

*Bandwidth Limitation:* Wang et al. [24] have shown that continuous streaming of High Definition (HD) video can saturate the bandwidth even with a small number of video sources. To analyze applications like counting cars (Query 2), it is not feasible to offload full video analytics to high-end nodes as it leads to higher bandwidth consumption. For example, complex event queries like counting car for high volume traffic monitoring [8] may happen less than 10% of the day. Transmitting the full video to cloud nodes stresses network bandwidth and increases latency.

### 2) Quality of Service Level Bounds

*Throughput and Latency*: Processing a continuous stream of video data on the cloud results in increased latency and reduced throughput as the system is also processing video which is not relevant to the user's queries. For example, Query 2 is only interested in 'Car' object and processing any non-relevant object frames affects resources and Quality of Service (QoS) of applications, especially those involved in real-time analytics.

*Accuracy*: Techniques like low-cost specialized DNN models [25, 26] have been proposed which can be deployed on the edge devices to improve video processing performance. However, low-cost models struggle to detect accurate events over videos with high occurrences of false positives and negatives.

### D. Proposed Approach

Processing IoMT data streams like videos pose significant challenges at edge nodes due to their unstructured and resource-intensive processing requirements. Current video-based CEP solutions [8, 27, 28] do not provide adaptive optimization over video data at the edge. Windows can play a crucial role in analyzing the content of incoming video data and can exploit video and DNN characteristics to improve system performance. The three key requirements to enable adaptivity for CEP windows over edge video streams are:

- Identify and exploit CEP *query characteristics* for videos to maintain edge resource usage (such as CPU and memory) and the quality of results (like accuracy*).
- Identify *optimal windows placement design* in CEP operator graph for fast *state-based* video event analytics (*high throughput*) and lower response time (*low latency*).
- Identify data mining techniques to *filter and transmit* video data to the cloud which have a high probability of events of interest to user queries using *resource-constrained* (CPU and memory) edge devices to *reduce network bandwidth.*

Thus, our goal is to create a *query and resource-aware windowing* approach to accelerate video event matching within resource-constrained environments. Following the concept of content-driven windows [21], an adaptive windowing technique VID-WIN is presented. VID-WIN adopts a query (such as accuracy, interested objects) and resource-aware (availability of CPU and memory) approach to exploit video content (like similar frames) and DNN properties to achieve an efficient QoS performance. We consider video event streams generated at the edge node where the underlying dataflow is composed of CEP operators that execute the video stream to perform pattern matching. As per requirements, the approach is discussed in detail.

### 1) CEP Query Characteristics

Event Query Languages are designed for the effective expression of complex events. Some of these query languages [8, 29] allow users to pass QoS parameters (such as accuracy, allowed false positive and negative rate) beyond standard query syntax for faster result with some error tolerance. Recently, the VidCEP framework proposed a novel Video Event Query Language (VEQL) [8] for video event detection. VEQL supports video event pattern matching using windows and accuracy QoS metrics in the CEP environment. This work extended the VEQL by adding two edge-based QoS metrics: `EDGE-CPU-USAGE` and `EDGE-MEMORY-USAGE` that allow users to limit edge resource usage with desired result accuracy. The work exploits the above discussed VEQL QoS metrics and query predicates (presence and absence of relevant objects) and proposes a query-driven windowing approach for faster video inference in an edge-cloud setting. In Fig. 2, two VEQL queries (Q1: Object Classification and Q2: Temporal Conjunction) with sliding windows and QoS metrics are defined. In the rest of the paper, these two queries will be evaluated to validate the proposed approach.

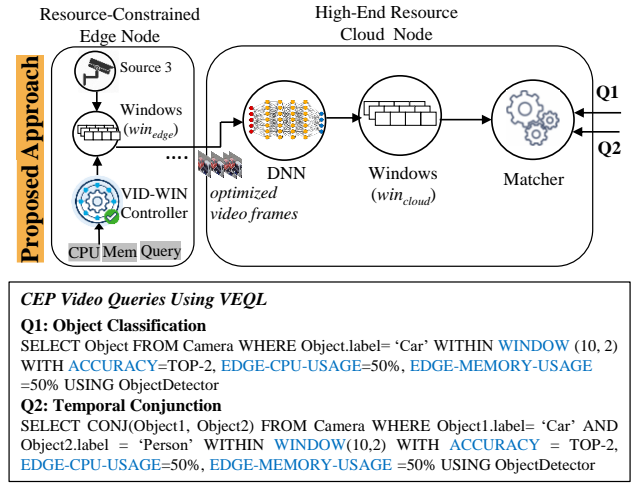

**CEP Video Queries Using VEQL**

**Q1: Object Classification**
SELECT Object FROM Camera WHERE Object.label = 'Car' WITHIN WINDOW (10, 2) WITH ACCURACY=TOP-2, EDGE-CPU-USAGE=50%, EDGE-MEMORY-USAGE =50% USING ObjectDetector

**Q2: Temporal Conjunction**
SELECT CONJ(Object1, Object2) FROM Camera WHERE Object1.label = 'Car' AND Object2.label = 'Person' WITHIN WINDOW(10,2) WITH ACCURACY = TOP-2, EDGE-CPU-USAGE=50%, EDGE-MEMORY-USAGE =50% USING ObjectDetector

Fig. 2. The paper presents a 2-stage windowing approach VID-WIN for video streams. VID-WIN windows run in parallel over the edge ($win_{edge}$) and cloud ($win_{cloud}$) nodes performing an optimized state-based video event matching by exploiting video content and DNN properties under given edge resource and query budget.

### 2) A 2-stage windows placement over edge and cloud node

As shown in Fig. 2, differing from the traditional approach of a single window (Fig.1), the proposed VID-WIN approach decomposes windowing into two stages and places them over edge ($win_{edge}$) and cloud ($win_{cloud}$). These two windows run in parallel over edge and cloud nodes, respectively for efficient state-based video event matching. The $win_{edge}$ constitutes a VID-WIN controller that dynamically adjusts performance at runtime. At edge, as a pre-processing step, VID-WIN controller uses CEP query predicates like objects, accuracy, and resource availability to send only relevant and optimized video frames to the cloud. These optimized video frames take less bandwidth and have relatively faster inference over DNN models (in the cloud), accelerating overall event matching performance. The proposed method emulates both the sliding and tumbling nature of windows with an optimization focus on video streams.

### 3) Adaptive Data Mining Techniques

The paper proposes four data mining approaches: *micro-batching*, *dynamic resizing*, *partial caching*, and *filtering* which enables VID-WIN to exploit low-level video content and DNN model properties for faster video inference. VID-WIN controller analyzes the video frames at $win_{edge}$ and creates dynamic *micro-batches* of similar frames. Later, a dynamic aspect ratio resizing technique is proposed that optimally *resizes* the micro-batch based on query accuracy. Two filtering methods *eager* and *lazy* are proposed to filter the frames from resized micro-batch. Filtering is based on a cache and window-based utility score derived using query objects and resource availability. The *micro-batch resizing* and *filtering* approach saves the bandwidth and have faster inference over the DNN model at the cloud. The key characteristic of VID-WIN approach is that it preserves window state while performing optimizations both at edge and cloud nodes.

The next section conceptualizes the formal background of windows operations, data mining approaches in the edge and introduces key concepts in DNN models for video analytics.

## III. BACKGROUND

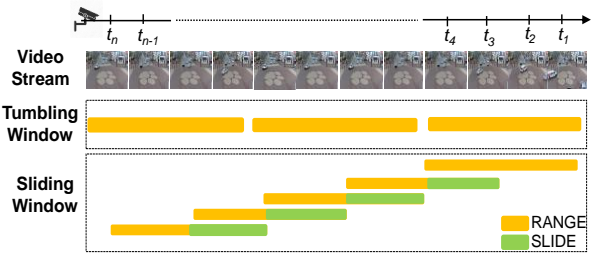

Fig. 3. Tumbling and sliding windows over video frames.

### A. Window- A Stateful Operator

Windows are stateful operators and apply computation over the sequence of input data. They act as an abstraction to discretize continuous data streams [9]. Windowing techniques like tumbling, sliding, and sampling of time and count categories support different query logic. Eq.1 shows a general window equation that depends on two parameters: *RANGE* and *SLIDE*. As shown in Fig. 3, *RANGE* computes the amount of a data stream that a window will ingest like $(t_1 - t_4)$ sec. *SLIDE* controls how much new and old data in the window can be consumed or discarded. The tumbling window always ingests new data items and discards the old data on completion. On the other hand, sliding windows keep both old and new data to avoid missing patterns but leads to the redundant computation.

$win = WINDOW(RANGE, SLIDE)$ where $RANGE$, $SLIDE$ $\epsilon$ [time, count]

$$where = \begin{cases} if\ RANGE > SLIDE & (SLIDING\ WINDOW) \\ if\ RANGE = SLIDE & (TUMBLING\ WINDOW) \\ if\ RANGE < SLIDE & (SAMPLING\ WINDOW) \end{cases} \quad (1)$$

$$S_{video} = \{(f_1, t_1), (f_2, t_2), \dots \dots \dots \} \quad (2)$$

$$win\ (RANGE, SLIDE)[time]\ (S_{video}) :\to S' \quad (3)$$

In Fig. 3, the windows capture the image frames from video streams. As per eq. 3, a window ($win$) is applied over an incoming video stream $S_{video}$ (eq. 2) and gives a fixed subsequence $S' = \{(f_1, t_1), (f_2, t_2), \dots \dots \dots (f_4, t_4)\}$ based on time. In eq. 2, time is taken as discrete for each frame ($f_i$) and arranged in a linear order $\{t_1, t_2, t_3 \dots \dots\}$ where $t_i < t_{i+1}$. This work focuses on time-based sliding and tumbling window.

### B. Data Mining for IoT and IoMT applications at the Edge

With the proliferation of sensors and IoT advancement, edge-based techniques are gaining wide importance that focuses on processing data near the source or at the network's periphery [30]. Processing data near to source improves the Quality of Service (QoS) such as reduced latency, less energy, smaller bandwidth usage, storage, and monetary cost savings. With the benefits, edge computing comes with challenges and require sustainable solutions for deployments. Different tailored edge-based data mining techniques like classification, time series analysis have been proposed for IoT scenario [31, 32]. Savaligo et al. [33] emphasized the data mining at the edge from three perspectives, data (heterogeneous in terms of format like audio, video, velocity like streaming data), device (memory, communication, energy features) and infrastructure. Recently, simulation-driven frameworks like EdgeMiningSim [31], EdgeCloudSim [33], iFogSim [34] and IOTSim [35] have been proposed to simulate edge-based IoT testbeds and benchmark data mining techniques.

Similarly, video-based IoMT applications may require a short response time under resource-constrained scenario for real-time event pattern matching. Different data mining techniques like background filtering [36], approximating model knobs [37], resource allocation [7], downsampling [14] and compression [38] have been presented for real-time video stream processing at the edge. In event processing, the data mining techniques have been proposed for traditional windows for scalar IoT data [21, 22] and lack mining methods to optimize video event matching which is the core of this work.

### C. Deep Neural Network for Video Analytics

Deep learning techniques like Convolutional Neural Networks (CNN) have become state-of-the-art methods in many computer vision tasks. A CNN consists of multiple stacked layers and extracts the image features as vector embeddings. A fully connected layer in a CNN with weights $W$ and bias $b$ has its activations represented as-

$$h^{(t)} = \sigma(W^{(t)T} h^{(t-1)}(x) + b^{(t)}) \quad (4)$$

In eq. 4, $h^{(0)}$ represents the input layer, so, $h^{(0)}(x) = x$ and $\sigma$ is a non-linear activation function. $W^{(t)}$ and $b^{(t)}$ are the weights and bias of the layer $h^{(t)}$. The function of a CNN layer is to capture the local spatial connectivity in the input where the size of the filter acts as the receptive field. Convolution filters are typically followed by Rectified Linear Unit (ReLU) activation and Pooling layers. The ReLU activation develops non-linear features, and the pooling operation extracts the highest activation or the average activation from the CNN filter output.

Pooling layers [39] are inserted between successive convolution layers to downsample and reduce the spatial dimensions by keeping the most important features or the representative features. Global pooling filter downsamples the entire input activations to a single value. It is usually used near the end of CNN as they dramatically downsample the output dimensions and more convolution filters after it will not provide meaningful output. CNN models like ResNet [41] and MobileNet [42] can perform object classification with reasonable accuracy. Similarly, YOLO [13] and Faster R-CNN [43] are CNN-based object detection models that can predict objects in the image with localized bounding boxes.

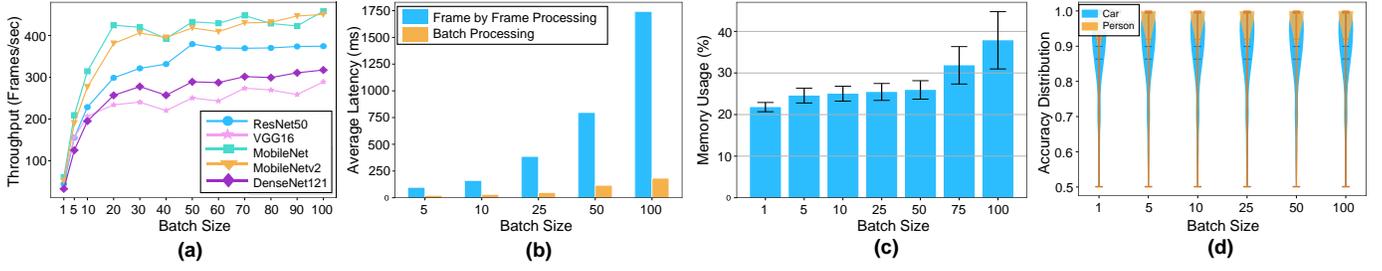

Fig. 4. Frame Batching: Performed on Nvidia RTX 2080Ti on Sandy Lane [40] video (a) Throughput (b) Latency (c) Memory usage (d) Accuracy distribution.

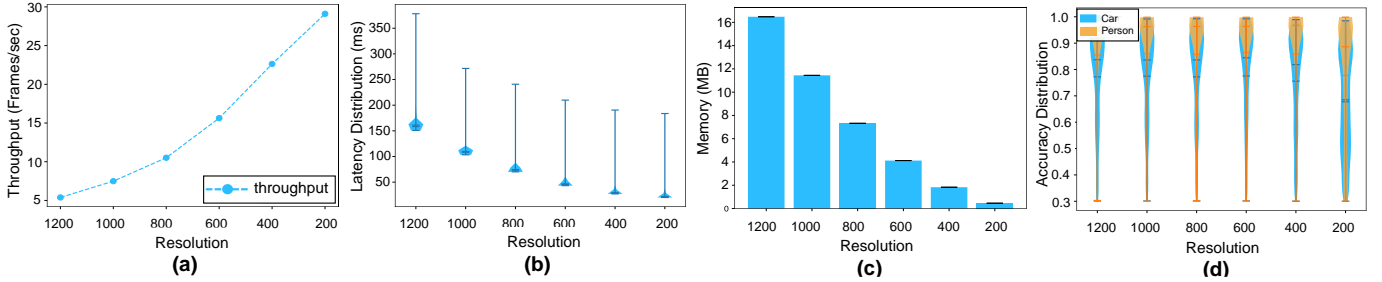

Fig. 5. Frame Resolution: Performed on Nvidia RTX 2080Ti on Sandy Lane [40] video (a) Throughput (b) Latency (c) Memory Usage (d) Accuracy distribution.

### D. Tunable Knobs in Deep Learning Models

TABLE I. METHODS FOR TUNING KNOBS IN DEEP LEARNING MODELS

| Input-Based | Model-Based | Sharing-Based |
|---|---|---|
| • Frame Rate<br>• Batch<br>• Frame Resolution<br>• Quantization<br>• Tiling<br>• Color Depth<br>• Region of Interest | • Matrix Factorization<br>• Matrix Pruning<br>• Architectural Changes<br>• Specialized CNNS<br>• Model Catalog<br>• Sliding Window Rate | • Sharing common layers<br>• Sharing common backbone models |

Based on the system and application-specific requirements, different parameters in deep learning models can be adjusted to optimize the performance. It is commonly known as tuning knobs in video analytics and can be primarily classified into three categories: 1) Input-based, 2) Model-based, and 3) Sharing-based. Input-based knobs try to transform the video data by changing frame rate and frame resolution [44, 45] or processing single channels image [46] which can impact DNN inference time. Model-based knobs focus on tuning deep learning models such as architectural changes by removing layers or creating specialized DNN models [47] for specific purposes. Sharing-based techniques emphasize sharing the layers or models [48, 49] for initial computation and then fanout output to application-specific models. Table I enlists different methods to tune knobs in deep learning models. Our work focuses on input-based knobs, which are applied over windows for incoming video streams to optimize the CEP performance.

## IV. INPUT BASED TUNABLE KNOBS OVER DNN MODELS

Inference time over DNN models is high. As discussed in Section III-D, different optimization techniques have been proposed for their faster execution. These techniques tune different model parameters as per the application objectives to optimize the model performance. Works including MCDNN [25] and NoScope [47] have proposed to create *specialized DNN models* to improve system performance. Since video data is highly dynamic and frequently changes (in seconds), loading different optimized model in memory will incur high runtime cost. Thus, we focus only on *input-based* transformation parameters where a model, once loaded, can accept different input configuration types from the windows. We focus on four crucial input parameters which have a significant impact on overall DNN model execution performance- 1) Batch Size, 2) Frame Resolution, 3) Frame Rate and 4) Region of Interest (ROI). Several experiments were performed to analyze the efficacy of input parameters on different system metrics.

### A. Batch Size

The DNN model's execution time is significantly lower if it receives the batch of input frames as a higher dimension tensor. Frame-level processing stalls the kernel memory leading to high processor utilization. Batching prevents the kernel from loading the input data every time and amortizes frame-level invocation overheads. The batch size significantly affects performance in terms of throughput and end-to-end latency with no change in prediction accuracy. Experimental results in Fig. 4 (a) shows the impact of different frame batch sizes over the pre-trained DNN models. The ResNet50 model throughput increases more than 8.7X from 40 frames per second (fps) to 350 fps when the batch size changes from 1 frame to 100 frames. MobileNet performs even better with a 9.6X throughput execution of 480 fps on a batch size of 100 frames. Fig. 4 (b) and (c) shows that the average latency of batch processing is 7X less with only 1.3X memory usage overhead compared to the frame-by-frame processing for the same number of frames. Fig 4 (d) shows the accuracy distribution of two objects with no change in accuracy across batches. It is evident from Fig. 4 that-'*batching of image frames improves model throughput performance and reduce latency as compared to frame level processing with minimal memory overhead and no change in accuracy*'.

### B. Frame Resolution

Frame resolution is another input transformation technique to accelerate the model inference speed [14, 44, 50]. Reducing the image size decreases the input information leading to fewer operations over the DNN model. This results in accuracy reduction as the model is relying on less information to predict the output. Fig. 5 (a), (b) and (c) shows that decreasing frame resolution can increase model throughput by 6X and reduce the latency and memory by 3X and 16X respectively for 200*200 image as compared to 1200*1200 resolution image. This improvement comes at a cost where the prediction accuracy drops with lower resolution images. Fig. 5 (d) shows the accuracy distribution of two objects ('car', 'person') using the Faster R-CNN [43] model. The median 'car' object accuracy drops from 0.85 to 0.68 for 1200*1200 and 200*200 resolution images, respectively and a similar trend follows for the 'person' object. The 200*200 resolution accuracy distribution detects objects with a low accuracy score. Therefore, we can say that '*resizing improves throughput, latency and memory usage but reduces accuracy*'.

### C. Frame Filtering

Continuous streaming of video data can overload network resources. Filtering or sampling is a common technique used to balance the supply by dropping data as per the resource demand. Works like FFS-VA [51] and Reducto [15] are some video analytics systems focused on frame-based filtering. As less data is received and processed by the CEP matcher to match events, there is a high probability of missing the pattern. Thus, '*filtering reduces bandwidth and resource usage but at the cost of accuracy.*'

### D. Region of Interest (ROI)

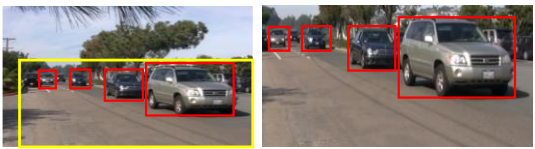

Fig. 6. Region of interest for 'car' object in a video.

In videos, objects exist in a specific region of a given frame. For example, car objects will usually exist on roads that are spatially located in one part of the video frame (Fig. 6). Processing the non-interesting region of an image where the probability of occurring objects is low wastes resources and affects the performance of an application. The ROI increases the overall detection accuracy as the DNN model does not process the non-interesting region (such as 'sky' and 'tree' in Fig. 6) which adds noise in the overall detection process.

Therefore, '*processing ROI achieves high accuracy and reduces the bandwidth usage*'.

Since the above four parameters significantly affect the model inference time, we utilize them as an adaptive factor in the windowing. In the next section, we discuss the impact of windows operator placement over the edge and cloud node and devise an adaptive windowing model for video event matching.

## V. VID-WIN: CONTENT AND QUERY-DRIVEN WINDOW OPERATOR FOR VIDEO EVENT ANALYTICS

This section discusses different window operator placement strategies for complex event processing. Later, it introduces the proposed adaptive windowing approach and describes the reference architecture.

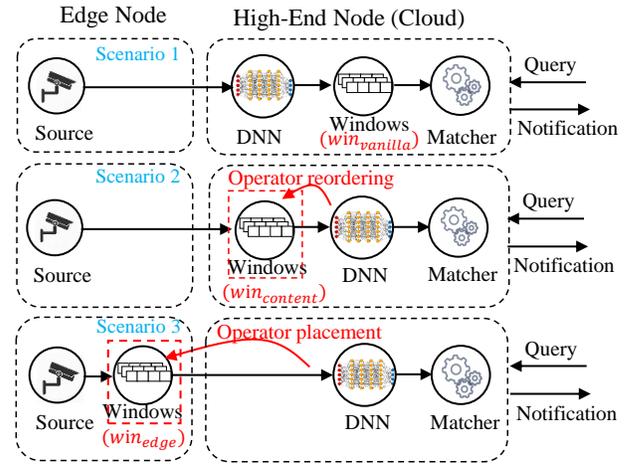

Fig. 7. (Top) Scenario1: A simple CEP operator graph deployed at the high-end node where windows operator is placed after DNN operator, (Mid) Scenario 2: Window operator is reordered and placed before DNN operator at the high-end node, (Bottom) Scenario 3: Window operator is placed at the edge node.

TABLE II. PERFORMANCE OF WINDOW OPTIMIZATION STRATEGIES FOR A VIDEO STREAM FROM AN EDGE (3GB, 4 CORES) TO HIGH-END NODE (RTX 2080 TI GPU, 16 CORES)

| Metrics | $win_{vanilla}$ | $win_{content}$ | $win_{edge}$ |
|---|---|---|---|
| Stream Optimization | None | Reordering | Placement |
| Strategy | Frame by frame processing, 1080p resolution | Batch size =10 frames, (800,800) resolution | Same as $win_{content}$ |
| Per Frame System Latency (ms) | 47.95 | 28.94 | **28.94** |
| System Throughput (fps) | 19.2 | 20.844 | **20.844** |
| Avg. Edge Memory Usage (%) | 13.23 | **13.23** | 23.2 |
| Avg. Edge CPU Usage (%) | 34.25 | 34.25 | **23.82** |
| Bandwidth Usage (MB/frame) | 6.3 | 6.3 | **1.95** |
| Avg. Accuracy | 0.94 | **0.94** | 0.92 |

### A. Windows optimization strategy over edge and cloud

Fig. 7 shows three window placement strategy across cloud and edge and their impact on the overall system performance.

*Scenario 1- Cloud only Vanilla Window:* In a general

streaming scenario, windows receive the structured data (such as temperature = 35°C) and then pass the stream state to the matcher for pattern mining. Fig. 7 (top) shows a vanilla window setting ($win_{vanilla}$). In the CEP operator graph, the window is placed after the DNN operator in a high-end node. As discussed in Section II-B, event processing is highly inefficient for video streams in $win_{vanilla}$ setting as it will miss the DNN-based optimization and impact the QoS results. Table II shows that in $win_{vanilla}$, the system achieves low throughput of 19.2 fps and a high per frame latency of 47.95 ms with increased usage of CPU and bandwidth.

*Scenario 2- Cloud only Content-driven Window:* Operator reordering [52] is a stream optimization technique where more selective operators are placed at the upstream of an operator graph to discard early data. Similar to adaptive windowing for structured data [21, 53], video content adaptivity can be equipped by placing windows ($win_{content}$) before costly DNN operators (Fig. 7 (mid)) and enabling input tuning knobs [54]. The $win_{content}$ placement can bring profitable gains to overall system performance. For example, Table II shows a $win_{content}$ operation where batching (10 frames) and low resolution (800*800) operation is enabled in windows with performance gains (such as low latency ~28.9 ms, high throughput ~20.9 fps) as compared to $win_{vanilla}$ strategy. The $win_{content}$ approach ensures *commutativity* [52]. It does not affect overall operator graph execution as the matcher will be receiving the processed frames of a given window state from the DNN operator. The key benefit and challenges in scenario 2 are-

- $win_{content}$ can tune input-based knobs of DNN (such as batching, resolution) which can result in *improved throughput* and *reduce latency*.
- $win_{content}$ strategy has some shortcomings as still full video data is transmitted from the edge to the high-end node with *no bandwidth savings*.
- Another drawback of $win_{content}$ is that it cannot perform window-based optimizations for reducing *redundant computations* of the sliding window. Pre-processing redundant video data will add another system overhead.

*Scenario 3- Content-driven window at Edge:* Operator placement [52, 55] is another stream optimization technique where operators are assigned to a specific host, cores or nodes. The key benefit of placement is to trade communication cost but with the assumption that the node on which the operator is deployed has enough resources to handle its functionality. Fig. 7 (bottom) shows a placement strategy where adaptive window operator ($win_{content}$) is placed on the edge node ($win_{edge}$). Fig. 8 shows general streaming operation where a HD video is livestreamed over the network from edge to a high-end node. The edge node struggles with high transmission latency and socket backpressure in frame-level processing as compared to batching (here 10 frames) scenario (Fig. 8 (b)) of $win_{edge}$. The benefits and challenges of $win_{edge}$ are:

- Scenario 3 is ideal in an edge-cloud setting where a data-driven operator ($win_{edge}$) is brought close to the edge.
- $win_{edge}$ can bring significant *bandwidth savings* as only required frames of specific resolution will be sent to the

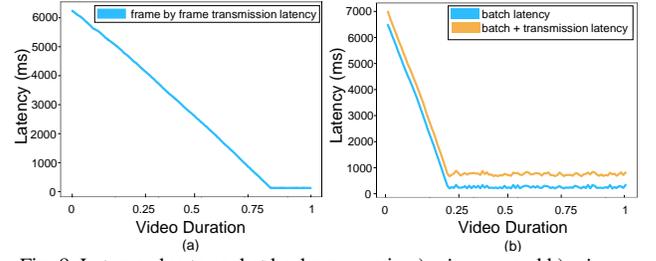

Fig. 8. Latency due to socket backpressure in a) $win_{content}$, and b) $win_{edge}$.

cloud. As per Table II, $win_{edge}$ requires a bandwidth of 1.95MB/frame as compared to $win_{content}$ which requires a bandwidth of 6.3MB/frame.

- Like scenario 2, $win_{edge}$ can *improve throughput* and *reduce latency* with minimal increase in edge memory usage (Table II).
- Since $win_{edge}$ transmits less data, it can bring significant cost saving for high-end nodes as their pricing includes storage cost.
- Similar to $win_{content}$, $win_{edge}$ will miss window-based optimizations like *redundant computations* in term of sliding windows.
- $win_{edge}$ can *overload* resource-constrained edge nodes with long sliding or tumbling windows.

Considering the above-discussed placement scenarios, a novel 2-stage windowing operator- VID-WIN is proposed for fast inference of video data. The following section focuses on the architecture and design of the VID-WIN method.

### B. VID-WIN: 2-Stage Allied Windowing on Edge and Cloud

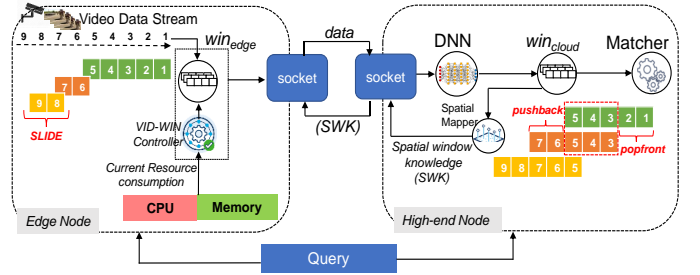

Fig. 9. VID-WIN high-level system architecture.

VID-WIN is a *2-stage allied window*, where two windows ($win_{edge}$, $win_{cloud}$) run in parallel across two nodes. VID-WIN optimizes DNN-based video stream inference between resource-constrained edge (low-end) nodes and high-end GPU enable nodes to perform state-based query matching. VID-WIN parameters are tuned adaptively based on *video content*, *resource budget* (CPU and memory) and the *query* to improve *application-level bounds* (latency, throughput and accuracy) and bandwidth usage. The 2-stages enable windows to perform both pre-processing and post-processing optimizations for fast video event analytics.

$$win_{edge} = \begin{cases} RANGE(t) & during\ initialization \\ SLIDE(t) & after\ initialization \end{cases} \quad (5)$$

$$win_{cloud} = \begin{cases} RANGE(t) & during\ initialization \\ \begin{pmatrix} pushback(SLIDE,t)[win_{cloud}] \\ popfront(SLIDE,t)[win_{cloud}] \end{pmatrix} & after\ initialization \end{cases}$$

Eq. 5 represents the windowing operation of VID-WIN over both nodes to manage the correct state. Fig. 9 shows a high-level system architecture of VID-WIN over edge and cloud. The $win_{edge}$ receives the video frames (like 1,2,3,4, ...) at the edge node and continuously transmits them to the cloud. The $win_{edge}$ consists of a *VID-WIN controller* which runs as a separate process and analyzes the incoming video content (similar to $win_{content}$) and performs optimization based on edge resource availability and query requirements. During the initialization phase, $win_{edge}$ works on the *RANGE* data and later switch to *SLIDE* mode (eq. 5). For example, in Fig. 9 a window of RANGE- 5 and SLIDE- 2 ($i.e.\ win(5,2)$) is defined where $win_{edge}$ will first perform the optimization process over RANGE data (i.e. 1,2,3,4,5) and later optimize only SLIDE data (such as (6,7) and (8,9)). The VID-WIN approach has several key benefits: 1) there is *no redundant computation* on the edge node as processing occurs only on *SLIDE* data, 2) *state-awareness* as $win_{edge}$ knows how much data needs to be ingested from the video stream, and 3) *content* and *resource-driven* optimization via *VID-WIN controller* to improve the application QoS. The *VID-WIN controller* design and adaptive strategy are discussed in detail in Section VI.

The $win_{cloud}$ placed on a high-end or cloud node runs in parallel with $win_{edge}$. The $win_{cloud}$ is placed after the DNN operator and receives the processed information (like objects). During initialization, the $win_{cloud}$ ingests only RANGE data (eq. 5). After initialization, $win_{cloud}$ applies *pushback* and *popfront* operations, respectively, to ingest new and discard the old SLIDE data. Fig. 9 shows that $win_{cloud}$ initially receives RANGE data (1,2,3,4,5) and applies *popfront* to discard old data (1,2) and *pushback* method to ingest new *SLIDE* data (6,7) maintaining the new window state (i.e. 3,4,5,6,7). The $win_{cloud}$ preserves the state even if $win_{edge}$ filters the video frames at the edge and processes only the information valid for that given state. For example, suppose in the above example $win_{edge}$ drops frame (5,6) then $win_{cloud}$ will only transfer state (3,4,7) to the matcher. The state is preserved as all the frames are time-stamped. Both windows ($win_{edge}, win_{cloud}$) run in parallel and keep track account of the time when the frame is received. In post-processing optimization, $win_{cloud}$ pass the processed DNN information to the *Spatial Mapper* operator that learns the Region Of Interest (ROI) of the queried object in the video. The ROI can later be passed to the VID-WIN controller to process only interesting areas in the frame. The concept of *Spatial Mapper* is similar to learning data characteristics over windows [21, 56] which can later be utilized to optimize the process further.

### C. Benefits of 2-Stage Allied Windowing

There are several benefits of VID-WIN as compared to other scenarios discussed in Section V-A. The $win_{edge}$ emulates Scenario 3, as it is placed at the edge with content adaptivity feature ($win_{content}$). Similarly, $win_{cloud}$ mimics Scenario 1 ($win_{vanilla}$) with the added functionality of data learning using the *Spatial Mapper* operator. Some of the key benefits of the 2-stage allied window approach are:

- Optimization can be performed both at the *pre-processing* and *post-processing* stage of streaming video data. The VID-WIN controller of $win_{edge}$ can perform pre-processing optimization by adaptively tuning input knob parameters like frame resizing, batching and filtering based on query and resource constraints. Similarly, the $win_{cloud}$ can optimize post-processing operation by finding the region of interests of query objects which can be utilized by $win_{edge}$ to further improve the system performance. The $win_{cloud}$ post-processing optimizations are not limited to objects spatial mapping and can be extended to other aggregation window optimizations [17-19, 57, 58] depending on the application context.
- The VID-WIN design applies to both tumbling and sliding windows which are the most widely adopted windows used across many streaming applications.
- Most of the video analytics optimization approaches are focused on frame-level object detection without any insight into complex event patterns. Recent work like FFS-VA [51], CloudSeg [14], FilterForward [49] and Reducto [15] act as *state-agnostic* filters to reduce video data without any state management and focus only on frame-based filtering. On the other hand, VID-WIN performs *state and query aware filtering* over video streams to detect complex patterns which span over time.

After discussing the benefits of 2-stage windowing over edge and cloud, the next section deep dives into the adaptive techniques and architecture which VID-WIN adopts for faster video inference.

## VI. VID-WIN ADAPTATION STRATEGY FOR INPUT KNOBS

This section conceptualizes the VID-WIN controller architecture at $win_{edge}$ and the data-driven techniques it uses to improve performance. One fundamental assumption is that the methods are applied over videos with a fixed background as most CCTV cameras have fixed Field of View (FoV).

### A. VID-WIN Controller Architecture

Fig. 10 shows a VID-WIN controller component that acts as the first-class abstraction over $win_{edge}$ to transform input frames for low-cost video inference. The numbers marked in the diagram represents the four stages of the VID-WIN controller which are discussed below:

1. *Create Micro-Batch:* Identify incoming frame sequences ($F1, F2, F3..$) over $win_{edge}$ by exploiting interframe similarities ($D1, D2, D3..$) using *Similarity Algorithm* and create an optimal micro-batch of the sequence of frames. The components *Similarity Score* and *Micro-Batch Size Analyzer* keeps a check on the size and create new micro-batches. The adaptive micro-batching strategy is discussed in detail in Section VI-B. The *Eager Micro-Batch Filter* component discards early frames to avoid further processing of frames when there is no change in the video content and is further discussed in Section VII.
2. *Resize Micro-Batch:* The *micro-batch resizer* resizes the resolution of micro-batch as per the *keyframe* resolution.

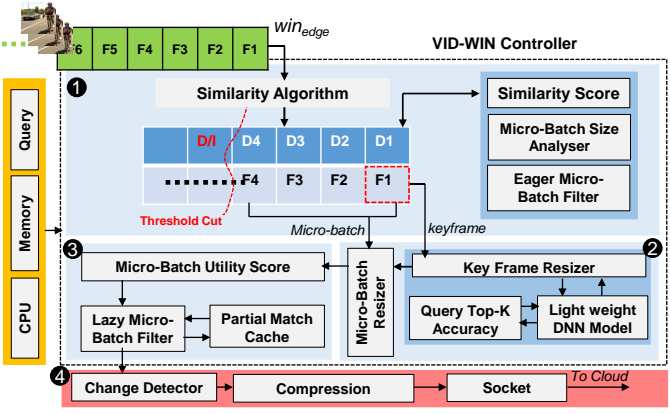

Fig. 10. VID-WIN controller architecture

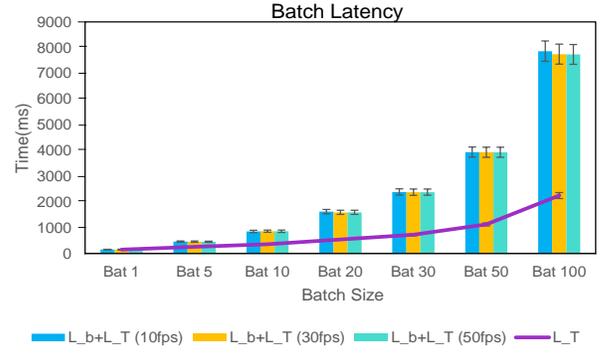

Fig. 11. Batching and transmission latency of different batch sizes for a 1080p Jackson Hole [59] video from the edge to cloud node.

The *KeyFrame Resizer* identifies the optimal resolution of a keyframe by passing it to a *Lightweight DNN model* and resize it as per the CEP query accuracy constraints. The query-based resizing strategy is discussed in Section VI-C.

3. *Filter Micro-Batch:* The *Micro-Batch Utility Scorer* identifies the *utility score* of the resized micro-batches. The *Lazy Micro-Batch Filter* can then discard an entire batch or set of frames from the batch depending on the utility score, resource availability (CPU and memory) and potential of getting a query pattern match (*Partial Match Cache*) over it. The filtering strategy is discussed to length in Section VII.

4. *Send micro-batch to the cloud:* The resized and filtered micro-batches are sent to the cloud node for processing. To further reduce bandwidth consumption, the *Change Detector* component sends only the *difference values* of the micro-batch with *Compression* to reduce the message payload over the network.

The next section conceptualizes the adaptive algorithm utilized by the above VID-WIN controller components.

### B. Dynamic Micro-Batching

As discussed in Section IV-A, the batching of input frames can improve the DNN model inference but impact the CPU and memory. Batching of frames can be defined in terms of time, count or memory size. Fig. 11 shows the batching ($l_B$) and transmission latency ($L_T$) of a Jackson Hole [59] HD video streamed at different fps speed. The total latency ($l_B + L_T$) increases from 1 sec (single frame) to 7.5 sec (batch of 100 frames). So, the question arises- a) *What is the optimal batch size? b) Should the batch size be fixed or dynamic?* We investigated the effect of dynamic batch sizes for video-based CEP matching and proposed a dynamic *micro-batching strategy* based on video *content* characteristics.

#### 1) Lightweight techniques to identify Micro-Batches

Frames in video sequences are correlated over temporal bound. We aim to identify batches of *highly similar frames* that can be further optimized later. This can be achieved using temporal video segmentation techniques which identify scene changes in video frames and divide them into meaningful batch-

-es [60]. To avoid additional overhead, two lightweight techniques have been used: 1) video encoding and 2) frame similarity score to identify batches in real-time.

*Inter and Intra-frame video encoding:* MPEG standard encodes video as a sequence of a group of pictures (GOP). H.264 encodes video pictures into three types of frames- I, P and B [61]. I-frames (intra-coded) are independently encoded while P (Predictive) and B (bi-directional predictive) frames are encoded with reference to other frames. I-frames have unique information as it contains data from the same frame. This encoding information is leveraged to create a micro-batch whenever a new I-frame arrives.

*Image Similarity Score:* Relying only on encoding information of frames is erroneous as in the real-world scene change is highly contextual. Thus, we also calculate the similarity score between frames to identify abrupt scene changes. Different frame similarity techniques have been proposed in the literature ranging from low-level features such as pixels, edges, corner, histogram, hashing and high-level visual descriptors like SIFT, SURF and CNN embeddings. The high-level descriptor techniques such as CNN based similarity methods are time-consuming and are not a good fit for latency-sensitive CEP applications.

Fig. 12 shows the latency distribution of different hashing and color histogram similarity techniques. The histogram-based similarity matching latency ranges between 1-2 milliseconds (ms) while Wavelet hashing latency distribution lies between 17-30ms. Thus, color histogram similarity is fast and efficient and can be deployed in real-time settings even for high-rate video streams such as 50fps where the delay between two frames is ~20 ms. The color histogram technique is applied, where the corresponding frames are converted to HSV space and correlation distance between two histograms $d_{sim}(H_1, H_2)$ is used to calculate the similarity score [62].

$$d_{sim}(H_1, H_2) = \frac{\sum_I (H_1(I) - \overline{H_1})(H_2(I) - \overline{H_2})}{\sqrt{\sum_I (H_1(I) - \overline{H_1})^2 \sum_I (H_2(I) - \overline{H_2})^2}}$$

$$and \ \overline{H_K} = \frac{1}{N} \sum_J H_K(J)$$

In the eq. 6, $H_1$ and $H_2$ are the histograms of two images to be compared and N is the number of histogram bins.

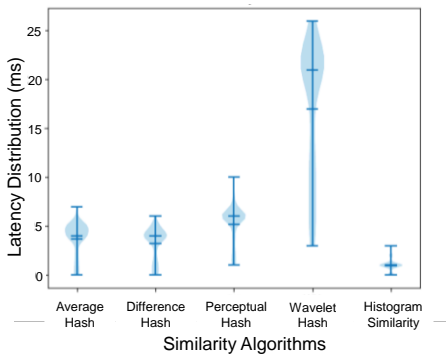

Fig. 12. Performance of different similarity techniques.

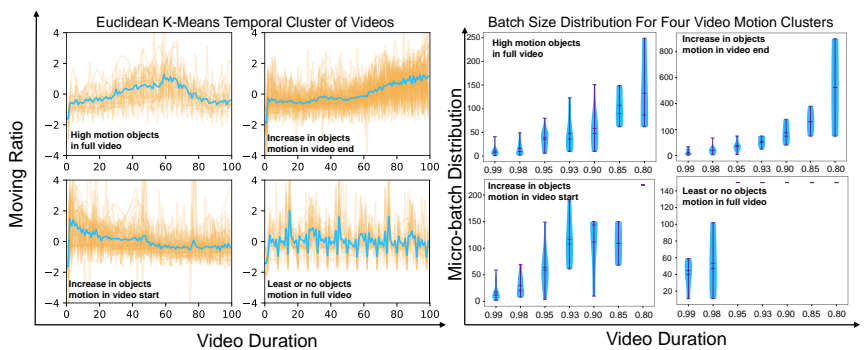

Fig. 13. Temporal clustering of videos as per different motion categories and its effect on micro-batch sizes for different similarity thresholds.

### 2) Identifying Similarity Threshold Score and Effective Batch Size

The histogram technique scores between 0-1 where a higher score is given for more similar images. The question arises what should be the similarity threshold score ($thres(d_{sim})$) to identify an optimal micro-batch size. The similarity across frames depends on the content, which is directly related to the object's motion. We applied a data-driven heuristic to determine the relationship between video motion characteristics and its effect on batch size distribution. A background subtraction technique [63] is used to identify the moving ratio of objects across frames. Fig. 13 (left) quantifies the different motion categories of objects across frames for different videos. A small dataset of 120 video clips (10-20 seconds each) was created based on different motion dynamics. The dataset is divided into four clusters using Timeseries Euclidean K-means clustering [64] having the following motion characteristics across time: 1) Slow Object Motion, 2) Increasing Object Motion, 3) Decreasing Object Motion, and 4) Continuous object motion. Fig. 13 (right) shows the batch size distribution of varying similarity scores for different motion clusters. The similarity score of 0.98 is set as the threshold score ($thres(d_{sim})$) as the median batch-size is around 35-60, which is an optimal size as the bigger sizes lead to higher latency.

There can be situations where there is no change in the video frames. This can lead to bigger batch sizes resulting in higher latency, memory, and CPU consumption. Offline profiling of batches is performed to understand the maximum optimal batch size ($MB_{max}$). Fig. 14 shows a 5-d plot of different batch sizes with throughput, latency, memory, and CPU. The three datasets are represented using colors with shade as memory and size as CPU consumption (solid color with bigger size means high memory and CPU consumption). The throughput and latency are measured till the processing of the DNN model in the cloud node, while CPU and memory correspond to the edge node. A costlier Faster R-CNN model is used for the profiling to get upper and lower bounds on latency and throughput. As per evaluation, the batch size of 70 keeps the right balance between high throughput and low latency with low CPU and memory usage. So, we fixed the max micro-batch size ($MB_{max}$) to 70 frames as a circuit breaker case when there is no motion or content change in the video stream.

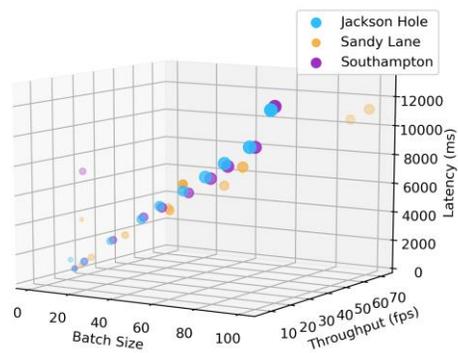

Fig. 14. Effect of batch size on throughput, latency, edge CPU and memory when different batches of frames are transmitted from edge to high-end node.

| **Algorithm 1: Adaptive Micro-Batching** |
|---|
| **Input:** *Incoming video frames* $\{f_1, f_2, f_3......\}$ *at* $win_{edge}$ *and SLIDE TIME of* $win_{edge}$ |
| **Output:** $Micro - Batch \ (MB)$ |
| 1: **for** *SLIDE TIME in* $win_{edge}$ **do** |
| 2:    **for** $f_i$ *in* $\{f_1, f_2, f_3......\}$ **do** |
| 3:       **if** $f_i = first \ frame$ **then** |
| 4:          $reference - frame \leftarrow f_i$ |
| 5:          $MB \leftarrow addtoMicroBatch(f_i)$ |
| 6:       **else** |
| 7:          **if** $f_i = I - frame$ **or** $d_{sim}(reference - frame, f_i) \leq thres(d_{sim})$ **or** $MB_{size} \geq MB_{max}$ **then** |
| 8:             $sendtoMicroBatchResizer(MB)$ |
| 9:             **GO TO** *Line* 2 |
| 10:          **else** |
| 11:             $MB \leftarrow addtoMicroBatch(f_i)$ |
| 12:    $sendtoMicroBatchResizer(MB)$ |
| 13: **GO TO** *Line* 1 |

Algorithm 1 defines the micro-batch strategy approach. The algorithm treats the first frame of the batch as a reference frame and calculates the similarity score with the reference frame. Whenever the $win_{edge}$ SLIDE ends, or it receives an I-frame, or the similarity score is less than the threshold score ($thres(d_{sim})$), or the batch size reaches to $MB_{max}$, the *Micro-Batch analyzer* creates a micro-batch and sends it to Micro-Batch Resizer. Thus, the SLIDE (during initialization its RANGE) of $win_{edge}$ (eq. 7) is a composition of several unique micro-batches ($MB_i$) for the video stream (S).

$$win_{edge}(S) = MB_1 \bigcup MB_2 \bigcup ... \bigcup MB_p \ , \forall MB_i \bigcap \forall MB_j \quad (7)$$

## C. Query-Aware Frame Resizing Policy

The rationale behind creating batches of similar images is to exploit the frame resolution knob and resize the whole batch to reduce memory, CPU, bandwidth usage, and DNN models execution time. One of the drawbacks of resizing is the loss in classification accuracy. Therefore, an effective strategy is required to resize the frames under given accuracy constraints. A novel query-aware frame resizing policy is devised which can be divided into three steps.

*Identifying Representative Frame in a Micro-Batch:* Video summarization techniques extract the keyframes from the given shot which is treated as a *representative or keyframe* for that whole video shot. To reduce overhead, we consider the first frame [65] of a micro-batch as our representative frame. All the similarity calculation for other frames in a batch is performed using this frame.

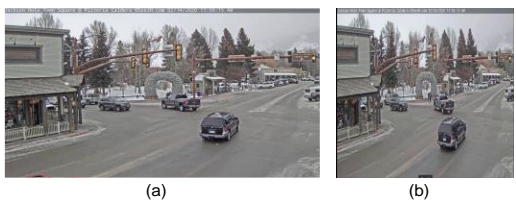

Fig. 15. Object shape distortion when (a) Aspect ratio is preserved and (b) Aspect ratio is not preserved.

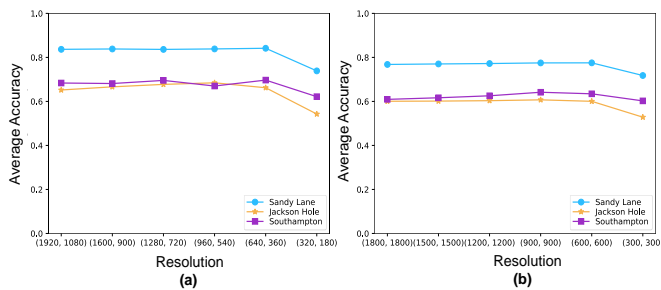

Fig. 16. Effect on accuracy with change in resolution for three datasets (a) Aspect ratio is preserved and (b) Aspect ratio is not preserved.

*Resizing Strategy:* In deep learning, it is common to use a square-shaped image resolution (e.g. 224*224) even though the CNN models are independent of image dimensions. The rationale behind using a square shape is that the model can precisely learn the features of images avoiding noises with better accuracy during training. 1080p, 4MP, 8MP are some of the standard resolutions of security and CCTV cameras with image sizes of 1920*1080, 2560*1440 and 3840*2160, respectively [66]. All the above resolution maintains an aspect ratio of 16:9. We propose to reduce the resolution of the image while *preserving the aspect ratio* of the image. Fig. 15 (a, b) shows that this approach maintains the object's shape with respect to the image, and the objects do not distort as may happen in square shapes (if aspect ratio is not preserved). Fig. 16 shows the accuracy of the frame resolution of three datasets with respect to square shape images where the accuracy of the aspect ratio preserving technique is ~13% to ~15% higher.

As per eq. 8, a resolution set ($R_s$) is created by reducing the size of image *height* ($H$) and *width* ($W$) with *aspect ratio* ($h:w$).

$$R_s = \{(H - h * 1, W - w * 1), (H - h * 2, W - w * 2)..... (h, w)\} \quad (8)$$

$R_s$ is combinatorically expensive as there can be multiple resolution sets. Therefore, a *candidate resolution set* ($CR_s$) is created by analyzing the resolution accuracy over our selected dataset. As shown in Fig. 16, there is significantly less or no change in accuracy across the higher resolution set. Thus, a candidate resolution set ($CR_s$) of following five resolutions [(288,162), (320,180), (480,270), (640,360), (960,540)] is selected while maintaining an aspect ratio of 16:9.

*Lightweight DNN classifier construction to accept the dynamic resolution of images:* Most of the pre-trained object classifier model (such as ResNet and MobileNet) accepts fixed resolution (like 224*224) of images. These models resize the image to a pre-determined resolution to classify the objects. The convolution layers are independent of image size, but the fully connected layer expects a fixed input size which hinders the model from accepting variable input size images. As discussed in Section III-C, the Global Average Pooling layer can reduce the feature map size from [$Batch(B)$, $Height(H)$, $Width(W)$, $Channels(C)$] to ($B,1,1, C$) making the output dimensions free of ($H$, $W$). MobileNet [42] is a lightweight image classification model with good performance for resource-constrained devices like smartphones. Work like Focus [67] suggests that even lightweight DNN model have high recall (nearly 100%) and detects the object in top-4 classes. We have created a lighter version of the MobileNet model by trimming its 50% layers and replacing the final layers with one GlobalAveragePooling layer followed by 3 Dense Layers and a 20 node softmax layer. This fine-tuned model can accept images of different resolution during runtime. The model was trained over Pascal VOC [68] dataset using a transfer learning approach.

*Identifying Optimal Resolution of Key Frame:* DNN models are probabilistic and give a score between 0 and 1 for each output. It is highly probable that even a high-end model predicts an object in the image with less score (such as 0.4). Therefore, instead of following an absolute accuracy score, we have followed the *top-k* approach to identify whether the model has predicted the object in the *top-k* position or not. For example, in Fig.2 query 1 (Q1), if the lightweight model can predict a 'Car' object in its *top-2* position, then the query pattern is satisfied. This information can be leveraged to resize the image to a minimum resolution where it can still predict information within the required top-k accuracy. As per VID-WIN architecture (Fig. 10), the representative frame (F1) is sent to the *Key Frame Resizer* where it predicts frame top-k accuracy using the *Lightweight DNN model* at a different resolution ($CR_s$) and sends a minimum frame resolution that satisfies the query accuracy constraints.

Since the lightweight DNN model is a multiclass classifier, there can be a possibility of score imbalance for required query objects. For example, query Q2 require two objects 'Car' and 'Person' to satisfy the conjunction (CONJ) pattern. It can be a possibility where both objects are present in the same frame. If both objects (such as 'Car':0.4, 'Person':0.3) are in the top-2 category (Q2 accuracy metric), then resizing happens until any object does not go out of the top-2 position. There can be a scenario where the score of objects may get skewed during resizing (such as 'Car':0.7, 'Person':0.003). To keep a check on such cases and maintain score accuracy intact during resizing, we check the $sum(top - 1 + top - 2 + .. top - k)(O) \geq 0 \cdot 45$ and ratio $\left(\frac{top-2}{top-1} .., \frac{top-k}{top-(k-1)} \geq 0 \cdot 45\right)$ accuracy score of identified query objects ($O$) and stop resizing if the above conditions are violated. In the above case, the value of $k$ (k=2 as per Fig 2) is based on the CEP query and the 0.45 score is empirically derived by running classification on multiple videos.

---

**Algorithm 2:** Adaptive Micro-Batch Resizing

**Input:** Candidate Resolution Set ($CR_s$), keyframe
        Query Objects ($O$), Query Accuracy ($top - k$)
        Lightweight DNN Model ($DNN$)

**Output**: keyframe optimal Resolution ($R_{optimal}$)

1:   $R_i \leftarrow startBinarySearch(CR_s)$
2:   **if** single object ($O$) in $DNN(keyframe(R_i))$ in top $- k$ **then**
3:     **repeat** Line 1 and 2 **until** object ($O$) **not** in top $- k$
4:     $sendtoMicroBatchResizer(keyframe, R_{optimal})$
5:   **if** multiple object ($O$) in $DNN(keyframe(R_i))$ in top $- k$ **then**
6:     **if** sumof top $- k$ ($O$) $> 0.45$ **and** $\left(\frac{top-2}{top-1} , ..., \frac{top-k}{top-(k-1)}\right) > 0.45$
7:       **repeat** Line 1,5,6 **until** objects ($O$) **not** in top $- k$
8:       $sendtoMicroBatchResizer(keyframe, R_{optimal})$
9:   **if** no object ($O$) in $DNN(keyframe(R_i))$ in top $- k$ **then**
10:   **repeat** Line 1 and 9 **until** max Resolution
11:   $sendtoMicroBatchResizer(keyframe, R_{min})$

---

Selecting different adaptations is based on the *candidate resolution set* ($CR_s$). The *binary search* is applied on $CR_s$ to bootstrap the search, and later resolution is chosen as per the previous keyframe resolution. The process restarts if the preceding keyframe lies in the second half of $CR_s$. In case if the object is not present in the keyframe, the lowest resolution image ($R_{min}$) is sent, as empirically the chances are high that the object is not present in that micro-batch. Algorithm 2 illustrates the proposed resizing policy. The *Micro-Batch Resizer* then resizes the full batch frames as per the received keyframe resolution ($R_{optimal}$). This adaptive *micro-batch resizing* of frames over $win_{edge}$ dramatically improves the overall system performance as shown in detail in the experiments.

## VII. Query Aware Adaptive Filtering under Resource Constraints

To reduce the stress of bandwidth and edge resources, VID-WIN uses a *query-aware caching* and micro-batch based *utility-driven resource-aware* filtering policy with the least impact on the quality of results. For a given $win_{edge}$, VID-WIN controller adopts a two-step filtering to drop frames at early and later stage of the processing depending on the nature of micro-batches.

### A. Eager Filtering

The eager filter is early-stage filtering performed at the micro-batching stage. The *Eager Micro-Batch Filter* (Fig. 10) drop the frames if there is no change in the video content for given consecutive micro-batches. For example, at night, the number of 'car' objects will be less, and there can exist long hours or minutes when no object is present. Transmitting such frames will only stress precious network and system resources. To avoid such situations, *Eager Micro-Batch Filter* continuously monitors the maximum micro-batch size

($MB_{max} = 70\ frames$) at $win_{edge}$. If there are consecutive $MB_{max}$ micro-batches indicating no activity happening, the eager filter will start dropping batches after the first $MB_{max}$ micro-batch transmission. The eager filter ensures that the first micro-batch of non-activity is not filtered and is forwarded to avoid potential pattern miss, as all other subsequent $MB_{max}$ are redundant batches. There is no query-awareness at this stage and filtering is based on low-level video content.

### B. Lazy Filtering

The *Lazy Micro-Batch Filter* discards resized micro-batches if they are not relevant to CEP query or violate available resource limits. It is divided into two parts:

#### 1) Partial Match based Cache Filtering over Edge

In CEP, a complex event pattern is made up of multiple simple or atomic events. For example, in Fig.2, query Q2 is interested in the complex pattern CONJ(Car, Person), which is made of two objects 'Car' and 'Person'. The CEP matcher performs pattern matching by correlating simple events (Car, Person) within a given window range to detect if a pattern exists or not. The minimal criteria for any video query ($Q$) to be fulfilled are the presence of all the query objects ($Q = \{o_1, o_2..., o_n\}$) in a window range. The object ($o_i$) can be treated as a *partial match* ($pm$) [69] if it is a subset of the query ($o_i \subset Q$) in a window. Thus, the micro-batch ($MB$) which do not consist of partial matches ($MB \nsubseteq Q$) can be filtered as it is not relevant for the CEP matcher during matching. For example, Q2 focuses only on objects 'Car' and 'Person' and any frame consisting of objects other than that can be discarded as it is not relevant to the query. In the current strategy, a complete (i.e., both 'Car' and 'Person' in the same frame) or partial match (i.e. only 'Car' or 'Person' in a frame) can exist in a micro-batch.

There can be multiple instances of event patterns within the same window. For example, for Q2 (CONJ(Car, Person)) in a video stream $[(Car, t_1), (Person, t_2), (Car, t_3), (Person, t_4)]$ there can be multiple event patterns like $[(Car, t_1), (Person, t_2)]$, $[(Car, t_1), (Person, t_4)]$ and $[(Car, t_3), (Person, t_4)]$. This combinatorically increases the cost of matching. To handle such situations, different selection (like first or last) and consumption strategies (such as consumed or zero) [70-72] have been proposed in CEP literature. In this work, we have used the *first selection* and *consumed consumption* policy for event matching. For the above example, the matcher will detect only two patterns, i.e. $[(Car, t_1), (Person, t_2)]$ and $[(Car, t_3), (Person, t_4)]$ for Q2 query and both patterns will not necessarily have the same accuracy.

Keeping edge constraints in mind, a new qualitative filtering policy is devised where the VID-WIN will forward only those objects whose accuracy score is higher than previous similar objects. This will lead to the detection of more qualitative event patterns reducing the overall resource consumption. A partial match ($pm$) *cache* is proposed based on the CEP query. Algorithm 3 illustrates the function of the cache. As the query is registered in the system, all the partial matches, i.e. objects are registered in the cache. All the values stored in the cache are only valid for the current state and the cache is re-initialized for every fresh window ($win_{edge}$). The cache stores the partial match accuracy value of a resized micro-batch obtained from a lightweight DNN model. The cache updates the incoming partial match value only when its accuracy (e.g. 0.50) is higher than previously-stored accuracy (e.g. 0.40) within the same window else it will drop the batch from further processing.

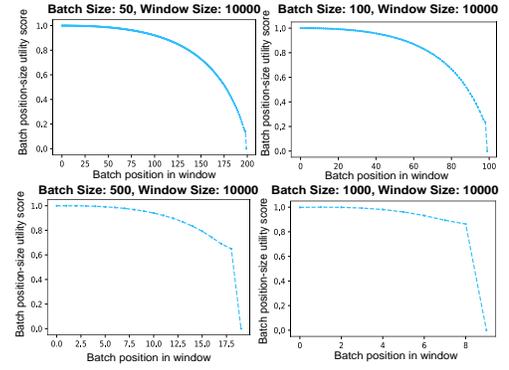

Fig. 17. Effect of micro-batch size and its relative position in windows over its utility.

-ar') exists earlier in the window then the probability of pattern matching ('Car' and 'Person') is high as still lot of frames are left to be processed. The relative position ($\omega_t$) of a micro-batch at a time (t) can be derived as-

$$\omega_t = \frac{win_{edge}(frame\ processed)}{win_{edge}(size)} \quad (11)$$

The $win_{edge}(size)$ is equal to the total number of frames a window can consume during a SLIDE (i.e. $slide - time \times fps$). Since the importance of micro-batch decreases with the progression of its relative position in the window, so the relative position importance ($rpi$) score will be:

$$MB_{rpi} = (1 - \omega_t) \quad (12)$$

For example, over a $win_{edge}(size)$ of 30 seconds (30*30fps = 900 frames), if the micro-batch (40 frames) occurs after 100 and 500 frames then $MB_{rpi}$ will be $\left(1 - \frac{100}{900}\right) = 0.88$ and $\left(1 - \frac{500}{900}\right) = 0.44$ respectively. It is evident from the above calculation that how the importance of micro-batch declines with its progression over the time window. The ratio of micro-batch size with the remaining window size is another essential factor to determine its utility. It is equivalent to-

$$MB_{win-remain} = 1 - \left(\frac{MB_{size}}{win_{edge}(remaining)}\right) \quad (13)$$

The probability of getting a partial match is high in larger sized micro-batches as compared to smaller batches. Using eq. 12 and 13, a combined utility score ($MB_{position-size}$) is formulated. $MB_{position-size}$ considers the micro-batch relative position importance ($MB_{rpi}$) and batch importance for the remaining window ($MB_{win-remain}$) using an entropy-based calculation which focuses on the average rate at which information is produced.

$$MB_{position-size} = \frac{\alpha + \beta}{(-log_2(MB_{rpi})) + (-log_2(MB_{win-remain}))} \quad (14)$$

$$where \quad \begin{array}{l} \alpha = MB_{rpi} * (-log_2(MB_{rpi})) \\ \beta = MB_{win-remain} * (-log_2(MB_{win-remain})) \end{array}$$

Fig. 17 shows the effect of micro-batch size and its relative position in windows over its utility. A bigger batch size at the starting of a window has a high position-size utility score as compared to the end of the window. Finally, the micro-batch utility score ($MB_{utility}$) is calculated as the entropy function like eq. 14 replacing values with $MB_{accuracy}$ and

---

**Input:** $Micro - Batch(MB_i), Micro - BatchObjectSet(MB_{object})$, $Query(Q), Cache(C)$
**Result:** $Forward\ Micro - Batch(MB_i), FilterMicro - Batch(MB_i)$,

1: **for** each window **do**
2:   $Initialize\ Cache(C) \leftarrow Q$
3:   **if** $MB_{object} \nsubseteq Q$ **then**
4:     $FilterMicro - Batch(MB_i)$
5:   **else**
6:     **for** all $o_i$ in $MB_{object}$ **do**
7:       **if** $o_i \subseteq Q$ and $Cache(C).isempty()$ **then**
8:         $Cache(C) \leftarrow updateCache(Accuracy(o_i))$
9:       **if** $o_i \subseteq Q$ and $Cache(C).notempty()$ **then**
10:        **if** any $Accuracy(o_i) > Cache(C)[Accuracy(o_i)]$ **then**
11:          $Cache(C) \leftarrow updateCache(Accuracy(o_i))$
12:          $Forward\ Micro - Batch(MB_i)$
13:        **else**
14:          $FilterMicro - Batch(MB_i)$

### 2) Resource-Based Micro-Batch Filtering

A novel *micro-batch utility score* ($MB_{utility}$) is devised where batches with low utility score can be discarded to keep the edge resources (memory and CPU) free. The micro-batch utility depends on two key factors: 1) The accuracy of the partial match ($pm$) present ($MB_{accuracy}$), and 2) The relative position and size of the batch in a window ($MB_{position-size}$). As per eq. 9 the $MB_{utility}$ is defined as the function of accuracy and batch position in the window.

$$MB_{utility} = U(MB_{accuracy}, MB_{position-size}) \quad (9)$$

#### a) Micro-Batch Accuracy

The micro-batch accuracy ($MB_{accuracy}$) represents the efficacy of the batch. It calculates the accuracy of partial matches (i.e. objects) if they are present in the query's top-k accuracy range (Q). The $MB_{accuracy}$ is calculated as-

$$MB_{accuracy} = \sum_{1=1}^{n} \frac{acu(o_i)}{(k)} \ where \quad \begin{cases} o_i \subseteq Q \\ k\ \epsilon\ top - k\ position \end{cases} \quad (10)$$

As per eq. 10, $MB_{accuracy}$ is the sum of the accuracy of objects $o_i$ which is a part of the query (Q) divided by its top-k position. For example, suppose the lightweight DNN model in the VID-WIN controller predicts the following objects with an accuracy of {'Car':0.4, 'Person': 0.3, 'Dog': 0.2. 'Bike':0.1} in a micro-batch. Then for Q2, the $MB_{accuracy}$ will be $\left(\frac{0.4}{1} + \frac{0.3}{2}\right) = 0.5$) as both 'Car' and 'Person' objects are present in the top-2 position as required in Q2.

#### b) Micro-Batch Relative Position and Size in Window

The CEP literature [73] suggests a strong correlation between pattern matching and the *relative position* of partial matches in a window. The probability of getting a pattern match is high at the beginning of the window which decreases gradually at the end of the window. For example, if a partial match (such as 'C-

$MB_{position-size}$. For example, for a given micro-batch if $MB_{accuracy} = 0.5$ and $MB_{position-size}= 0.20$, then $MB_{utility}$ will be 0.866. It means ~86% information of micro-batch is valuable to keep its quality maintained. The micro-batch utility ($MB_{utility}$) is then used to filter frames to a tradeoff with available edge resources.

---

**Algorithm 4: Dual Bound Resource Aware Filtering at Edge**

**Input:** $Micro-Batch(MB_i)$, $Query(Q)$, $MB_{utility}$

**Result:** $Forward\ Micro-Batch(MB_i)$
$FilterFrameinMicro-Batch(MB_i)$

1: **for** each $(MB_i)$ in window **do**
2:    **if** $M_{util} \leq Q_{mem-req}(M_{max})$ **or** $CPU_{util} \leq Q_{CPU-req}(C_{max})$
3:      $Forward\ Micro-Batch(MB_i)$
4:    **else**
5:      $FilterFrameinMicro-Batch(MB_i)$ **using**
6:      **do** $Subtract(M_{util}, M(MB_i) * MB_{utility})$ **or**
       $Subtract(CPU_{util}, CPU(MB_i) * MB_{utility})$ **until**
7:      $M_{util} \leq Q_{mem-req}(M_{max})$ **and** $CPU_{util} \leq Q_{CPU-req}(C_{max})$

---

A dual bound (memory and CPU) load filtering is explained in Algorithm 4. The filter drops frames from micro-batch using $MB_{utility}$ if the given query-based memory or CPU bound is violated. For example, Q2 in Fig. 2 puts a bound of 50% on memory ($Q_{mem-req}$) and CPU ($Q_{CPU-req}$) usage on maximum available resources (memory: $M_{max}$, CPU: $C_{max}$) of an edge node. If the above bounds are violated, then the filter will start dropping frames from the micro-batch. The maximum frames which can be dropped from a micro-batch depend on the $MB_{utility}$. For example, if $MB_{utility}$ is 0.86 then a minimum 86% of the micro-batch information will be kept. The remaining available frames can be dropped until CPU and memory consumption comes under the required resource bound (i.e. 50% in Q2).

Thus, the overall process is to create a *micro-batch*, *resize* them based on query accuracy, and *filter* the frames as per video content, partial query matches and resource bounds. The $win_{edge}$ does not hold any micro-batches at the edge and continuously streams the filtered and resized micro-batches to the high-end node. To further reduce bandwidth consumption, the compressed difference value of the micro-batches (using *Change Detector* and *Compression* component (Fig. 10)) is sent via socket. This is done by keeping the original keyframe and taking the difference of all the other frames in the micro-batch with respect to the keyframe. The batches are recreated at the high-end node using the keyframe. The $win_{cloud}$ performs post-processing optimization where the region of interest (ROI) of query-related objects are mapped using a Spatial-Mapper. This ROI can later be passed to the VID-WIN controller to crop the video frames further to focus only on interesting regions of the video.

## VIII. EXPERIMENT AND RESULTS

### A. Implementation and Datasets

The VID-WIN prototype[1] is implemented in Python 3. A

---

Docker container is used to simulate the edge device with different memory (500MB- 3GB) and CPU cores (1-5) settings. The VID-WIN controller is implemented using Python's multiprocessing functionality where all the processes run in parallel. FFmpeg and OpenCV are used for video streaming and pre-processing. The lightweight DNN model (MobileNet lighter version) in the VID-WIN controller is implemented using the TensorFlow Keras API, and its CPU version is deployed at runtime. The video dataset is mounted as *volume* over the Docker container to simulate streaming behavior. Docker stats is used for logging the container usage and network I/O.

The high-end node container is deployed on a Linux machine running with a 3.1 GHz processor, 64 GB RAM and Nvidia RTX 2080 Ti GPU. It runs the VidCEP [8] complex event matcher, Faster R-CNN [43] and ResNet101 object detection and classification model, respectively. The ResNet101 model is re-trained on the Pascal VOC dataset using a transfer learning approach to accept the different resolution of images. Brokerless messaging and publish-subscribe architecture is followed where data from edge to high-end node is transmitted via ZeroMQ [74]. The ZeroMQ sockets send discrete messages as compared to conventional sockets streams of bytes and is suitable for our purpose. The benchmarking with Esper CEP windows is implemented in Java with Deeplearning4j as the backend library to run the DNN model. The current experiments are focused on the performance of $win_{edge}$ as all optimization techniques are based on it. The $win_{cloud}$ currently handle only ROI knobs and will be mentioned explicitly when involved in the experiment. Table III shows the list of 5 datasets that are selected to validate the VID-WIN approach.

TABLE III. DATASETS

| Dataset | Resolution | Objects |
|---|---|---|
| Jackson Hole [59] | 1080p | Car, Person |
| Southampton [75] | 1080p | Car |
| Auburn Toomer's Corner [76] | 1080p | Person, Car |
| Sandy Lane [40] | 1080p | Person, Car |
| Times Square [77] | 1080p | Person, Car |

### B. Evaluation

#### 1) Effect of dynamic micro-batch vs fixed batch size over throughput and latency (static resolution)

The dynamic micro-batching performance is compared with the fixed batch (size of 1, 5, 10, 25, 50, 100) approaches having a static resolution. The batch size of 1 represents the frame-by-frame processing, and the resolution is fixed at 500*500. Since the comparison is about fixed batching vs dynamic micro batching, so the effect of window slide and range is not considered here. A ResNet101 [41] model trained on Pascal VOC [68] dataset is used for this experiment. The experimental values are logged when the system stabilizes with continuous transmission of data. Fig. 18 shows the throughput and latency performance of micro-batching over four datasets- Southampton, Jackson Hole, Sandy Lane and Auburn. As per eq. 15, latency ($L$) is measured as the time of reading and batch-

---

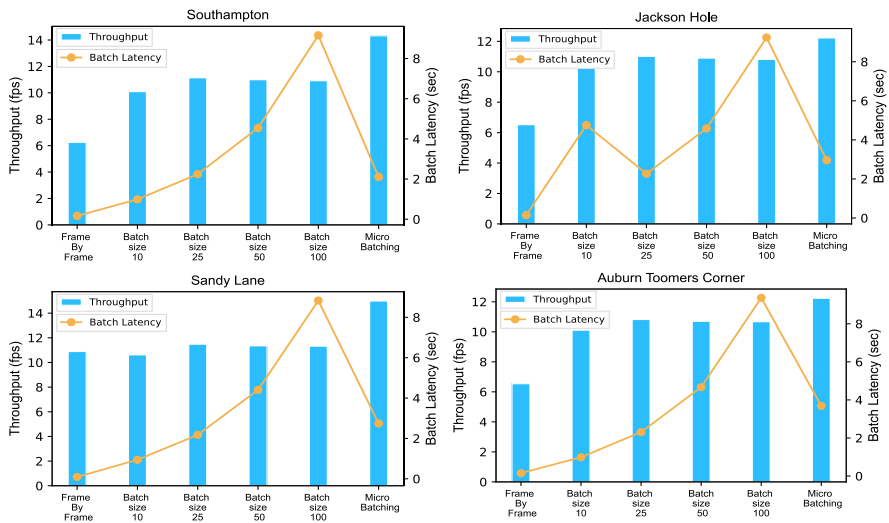

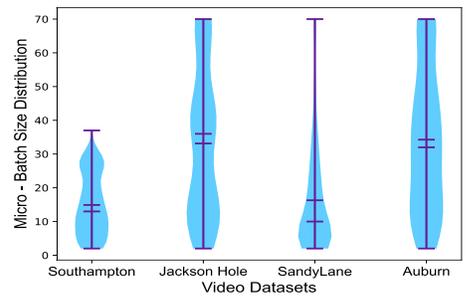

Fig. 19. Micro-batch size distribution over four datasets.

Fig. 18. Performance of dynamic micro-batching (fixed resolution) vs fixed batches over throughput and latency over four datasets.

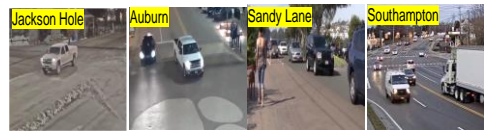

Fig. 20. Number of objects comparison over four datasets.

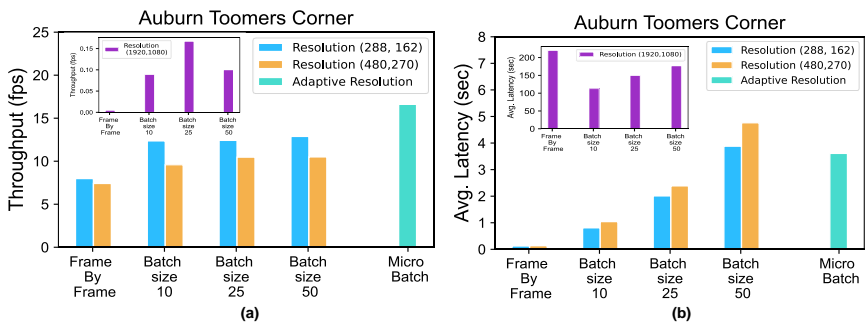

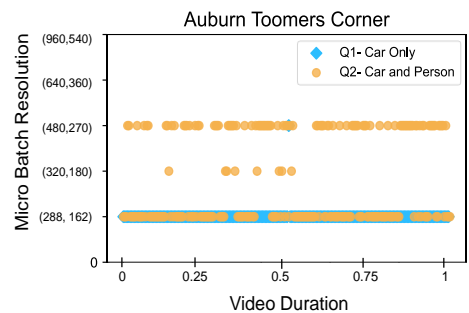

Fig. 21. Performance of dynamic micro-batch resizing vs batch of fixed resolution on (a) Throughput and (b) Latency for Auburn Toomer's dataset.

Fig. 22. Adaptive resolution selection for two queries over Auburn Toomer's dataset.

-ing frames at the edge node ($l_B$), transmission time to cloud node ($L_T$) and the DNN model processing time ($l_{DNN}$).

$$Latency \ (L) = l_b + l_T + l_{DNN} \qquad (15)$$

Since micro-batching includes dynamic batch sizes, the weighted mean is used to calculate the batch latency ($L_{Batch}$).

$$L_B = \frac{\sum_{i=1}^{n} L_i B_i}{\Sigma B_i} \qquad (16)$$

In the eq. 16, $B_i \ \epsilon \ batch \ size$. The throughput is the number of frames processed per second. The micro-batches throughput is calculated by logging the experimentation time with the number of frames being processed in that duration. As per Fig. 18, the rate of change of throughput is 30% to 40% higher for micro-batching as compared to all other fixed batch sizes. The micro-batching throughput of Southampton is 14.3 fps which is 2.3X and 1.3X higher than frame by frame (6.25 fps) and 50 batch (10.92 fps) processing. The same throughput pattern is followed for all other three datasets. The batch latency ($L_B$) increases with the number of frames in a batch. The micro-batching strategy outperforms static batches in terms of batch latency. The batch latency of micro-batch in the Southampton dataset is 2.1 sec and is 4.3X less than for 100 batch (9.15 sec) and 1.2X less for 25 batch (2.44 sec) sizes. A similar batch latency pattern is followed across all four datasets. The micro-batching batch latency is higher than frame by frame processing and other smaller batches as they process fewer frames resulting in lower throughput.

Fig. 19 shows the distribution of micro-batches of all four datasets. The median batch size of Jackson Hole and Auburn ranges between 35-38 frames while that of Southampton and Sandy Lane ranges between 11-15 frames per batch. This gives an interesting fact regarding the nature of datasets regarding the number of objects and their motion dynamics. Jackson Hole and Auburn have fewer objects with less motion leading to bigger batch sizes compared to Southampton and Sandy Lane where the number of objects are very high (Fig. 20).

*2) Effect of VID-WIN micro-batch resizing vs batch size of different resolution over throughput, latency and query accuracy*

Micro-batch resizing further optimizes the overall system performance. Fig. 21 shows the comparison of batch sizes of different resolution with the micro-batch resizing technique for Auburn Toomer's Corner dataset for query Q1 (detect 'car' with top-2 accuracy). The batch size of 1,10, 25 and 50 are selected with the highest, mid and lowest resolution of (1920,1080), (480,270) and (288,162). The above resolution is selected to have a fair comparison as the micro-batch candidate resolution set ($CR_s$) has a medium and lowest resolution of (480,270) and (288,162) respectively. The original HD resolution (1920,1080) is considered to get the baseline performance. The micro-batch resizing technique achieves the highest throughput of 16.7 fps and is 2.11X and 1.4X higher as compared to the frame-by-frame processing and a batch size of 10 with the lowest resolution (Fig. 21 (a)). The higher throughput is due to the resizing of frames to the lowest resolution as the lightweight classifier is able to detect the 'car' object in the top-2 range as required by the query Q1. Fig. 21 (b) shows that batch latency of micro-batch resizing is 3.61 sec which is less than the lowest resolution of a batch size of 50 frames. Although the micro-batch resizing latency is higher with a batch size of 1 and 10, they achieve low throughput compared to our approach. The HD (1920,1080) streaming resulted in the worst throughput performance (sub bar graph in Fig. 21) with a throughput of 0.1fps and latency of 177 sec for a batch size of 50 frames.

Fig. 22 shows the query awareness and adaptivity of the proposed resizing policy over the Auburn video. The efficacy of the resizing is tested on query Q1 (detect 'car') and Q2 (detect 'car' and 'person'). The CONJ clause of Q2 is not evaluated here as it is a part of the CEP matcher. Here the emphasis is on how the resizing behaves with the different number of query objects. It is evident that in Q1, the resizer streamed most of the frames at the lowest resolution (288,162). The number of objects increases in Q2 ('car' and 'person'), so the resizer searches the optimal resolution where both objects can be found for required query accuracy (top-2). Thus, the resizer selects different resolution ((480,270), (320,180) and (288,162)) from candidate resolution set ($CR_s$) at various stages of the video for Q2 query.

TABLE IV. ACCURACY PERFORMANCE OF MICRO-BATCH RESIZING WITH DIFFERENT RESOLUTION

| Datasets | Resolution (288,162) | Resolution (480,270) | Resolution (1920,1080) | Micro Batch Resizing |
|---|---|---|---|---|
| Auburn | 0.688 | 0.701 | 0.703 | 0.679 |
| Southampton | 0.575 | 0.619 | 0.652 | 0.615 |

Table IV shows accuracy for the Auburn and Southampton dataset. Here the accuracy is the average prediction score given by the DNN model for each batch. Since accuracy is *batch independent* only resolutions are mentioned in the table. The average accuracy score of micro-batch resizing is 0.679 and 0.615 for Auburn and Southampton. The accuracy is only 0.024 and 0.037 less than the highest resolution score (1920,1080) for both datasets. Thus, the micro-batch resizing strategy is highly efficient, resulting in increased throughput and low latency with significantly less impact on the quality of the query result.

### 3) Evaluation of VID-WIN filtering policy on edge CPU and memory usage

*Memory Requirement for VID-WIN:* Table V shows the minimum memory requirement for deploying standard windows of a given size for streaming a 540p resolution video over an edge device. An average 540p video frame is around

~0.67 MB and requires approximately 20.1 MB memory space for handling a 1sec video stream (30 fps). As per Table V, edge devices cannot hold bigger window sizes (such as 30 min) because of the high memory requirement (35.33GB). On the other hand, VID-WIN adopts a different strategy where it continuously streams the micro-batches of frames while preserving the stream's state. VID-WIN requires a maximum of ~81.5MB of memory and is independent of window size. This is because VID-WIN streams a max micro-batch size ($MB_{max}$) of 70 frames which requires ~47 MB memory including other processing overheads.

TABLE V. VID-WIN VS NORMAL WINDOW MEMORY REQUIREMENTS OVER EDGE NODE

| Window SLIDE Size | Normal Window | | $win_{edge}$ | |
|---|---|---|---|---|
| | Memory required | Edge compatibility | Memory required | Edge compatibility |
| 10sec | 201 MB | Yes | 81.5 MB | Yes |
| 1min | 1.17 GB | Yes | 81.5 MB | Yes |
| 30min | 35.33 GB | No | 81.5 MB | Yes |
| 1hour | 70.66 GB | No | 81.5 MB | Yes |
| 10hour | 706.6 GB | No | 81.5 MB | Yes |

*Performance of VID-WIN micro-batch filtering at the edge:* Fig. 23 shows VID-WIN filtering techniques for Q1 (i.e. detect 'car') over Auburn Toomer's dataset across different window sizes. The SLIDE of the window ($win_{edge}$) size is selected as this is where the whole VID-WIN adaptive technique is focused. This selection is also valid for tumbling windows if the SLIDE length is equal to the window RANGE. Three filtering approaches are benchmarked to show the fractions of frame filtered-1) eager filtering, 2) eager plus cache filtering (query awareness), and 3) eager plus cache plus utility (resource awareness) on an edge device having 2GB RAM and 4-cores CPU. The utility filter is set at two resource bounds of 80% and 50% of the usage of CPU and memory. If the edge resource usage goes above a given bound, then the utility filter will drop more frames from the micro-batch to reduce resource consumption.

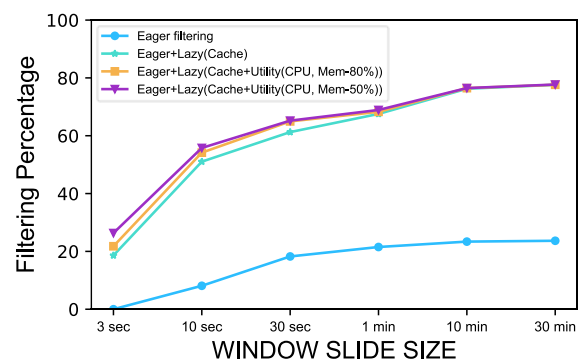

Fig. 23. Performance of different VID-WIN filtering techniques at the edge node on Auburn Toomer's dataset.

As per Fig. 23, the eager filtering percentage increases with the increase in window size. This is because the probability of repetition of consecutive max micro-batch size ($MB_{max}$) is less with smaller windows as the window closes early and the filter works afresh with the new window. With a window size of 30 min, the eager filter has dropped ~ 20% frames indicating no change in content during that time. Cache filters are the most effective filters as they only include frames having potential partial matches. Cache filters even work well for smaller windows and have filtered ~58% of the frames for a window size of 30 sec. One of the key reason for such aggressive filtering is that VID-WIN only forwards the partial matches which have a higher probability than previous partial matches available in the cache. As discussed earlier, VID-WINs overall memory usage is only ~81.5 MB. Thus, the utility filter drops significantly fewer numbers of frames. Most of the filtering occurs due to increased CPU usage instead of memory. It can be seen in Fig. 23 that the overall filtering with 50% memory and CPU usage is ~67% for a window of 30 sec and increases up to ~78.6% for a window of 30 min. The utility filtering at 50% resource usage is slightly higher than 80% due to more rigorous constraints on resource bounds. The utility filtering converges with cache filtering for long windows as resource usage stabilizes which is not in the case of smaller size windows.

TABLE VI. VID-WIN EVENT ACCURACY FOR QUERY Q1 AND Q2 ON TIMES SQUARE VIDEO

| Technique | Q1 (Car) | Q2 CONJ(Car, person) |
|---|---|---|
| VID-WIN+No filtering | 0.978 | 0.966 |
| VID-WIN+ filtering (eager) | 0.973 | 0.959 |
| VID-WIN+ filtering(eager+lazy(cache)) | 0.96 | 0.943 |
| VID-WIN+ filtering(eager+lazy+utility(80%)) | 0.953 | 0.940 |
| VID-WIN+ filtering(eager+lazy+utility(50%)) | 0.946 | 0.936 |

*Effect of filtering over event accuracy:* As discussed in the initial sections, filtering drops the overall query accuracy. A new event-centric query accuracy is used to evaluate such scenarios. The Q1 and Q2 are window-based temporal queries. For example, query Q2 (CONJ operator) pattern can be satisfied if the objects ('car', 'person') appears in a single frame or across the frames within the same window. Thus, the event existence can span across the frames and multiple events can be detected over the same window. Due to the temporal and event-centric query nature, detection of the single event will also satisfy the query. We have utilized the metrics devised in [49, 78] to compute the event accuracy. For a given query Q, the event accuracy is calculated as:

$$Event\ Occurence\ (EO) = \begin{cases} 1 & if\ an\ event\ is\ detected\ in\ win_{cloud} \\ 0 & if\ no\ event\ is\ detected\ in\ win_{cloud} \end{cases}$$

$$Event\ Extra\ (EX) = \frac{|E| \cap |G|}{|G|}$$

$$where \begin{cases} E\ \epsilon\ extra\ events\ detected\ in\ win_{cloud} \\ G\ \epsilon\ groundtruth\ events\ in\ win_{cloud} \end{cases} \quad (17)$$

$$Event\ Accuracy = \alpha * EO + \beta * EX$$

To give greater emphasis on event occurrence($EO$), the value of $\alpha$ and $\beta$ is set as 0.9 and 0.1 (eq. 17), respectively [49]. For example, if the ground truth events ($G$) is 10 and total events occurred is 5 then $EO = 1$ and EX = 4. The event accuracy will be 0.9*1+ 0.1*(4/9) = 0.94. To simplify the calculation a tumbling window of 3 sec is used to create the ground truth dataset. Table VI shows the event accuracy of Q1 and Q2 for the Times Square dataset for different VID-WIN techniques. The vanilla VID-WIN approach without filtering has an average event accuracy of 0.978 and 0.966 for Q1 and Q2. With filtering (eager + lazy (cache)), there is a minor drop of ~1.4% in the accuracy and is 0.96 and 0.943 for Q1, Q2, respectively. The utility filtering has a minimal effect on accuracy for both the queries. The Q2 accuracy is lower as compared to Q1 due to misclassifications from the DNN model where some car objects were detected as 'truck'.

TABLE VII. VID-WIN FILTERING COMPARISON WITH REDUCTO [15] ON SOUTHAMPTON DATASET

| Reducto | Frame Level Accuracy | Filtering Percentage |
|---|---|---|
| Reducto Optimal | 0.90 | 34.31 |
| | 0.85 | 55.71 |
| | 0.80 | 69.02 |
| | 0.75 | 77.41 |
| | 0.70 | 80.01 |
| VID-WIN | Window-based event accuracy | Filtering Percentage |
| VID-WIN + filtering (eager) | 0.969 | 2.11 |
| VID-WIN + filtering(eager+lazy(cache)) | 0.952 | 50.82 |
| VID-WIN+ filtering(eager+lazy+utility(80%)) | 0.947 | 55.95 |
| VID-WIN+ filtering(eager+lazy+utility(50%)) | 0.939 | 60.02 |

*Filtering comparison with Reducto:* Reducto [13] performs frame-level filtering based on object accuracy. Q1 is used as a baseline for Reducto comparison on the Southampton dataset. Reducto is initialized with its prebuilt threshold file using edge features, and its filtering percentage is benchmarked across multiple query accuracies (mAP-2) using YOLO [13] model. Table VII shows that Reducto achieves filtering of 34.31% for 0.90 query accuracy which increases up to 80.01% for 0.70 accuracy. VID-WIN filtering performs 60.02% filtering for a window size of 5 sec. VID-WIN eager filter achieves very low filtering of 2.11% due to the nature of video where there is a continuous motion of 'car' objects. The window-based event accuracy is greater than ~0.93 and is higher as compared to Reducto. There cannot be a direct accuracy comparison between the two techniques as Reducto uses frame-level filtering using high-end YOLO. While VID-WIN uses a lightweight DNN model and performs state-based filtering.

### 4) VID-WIN bandwidth saving comparison with the state-of-the-art edge-cloud vision technique

CloudSeg [14] is an edge to cloud vision analytics framework with a key focus on bandwidth savings. CloudSeg streams low-resolution frames from edge devices to save bandwidth and then apply super-resolution technique to upsample a high-resolution image on the cloud node. VID-WIN performance is compared with CloudSeg for bandwidth savings and other system metrics. The CloudSeg super-resolution model CARN [79] is embedded at the high-end node before the DNN model (ResNet101 in our case). We have followed a similar strategy as devised in Reducto and streamed video frames at 4X and 8X low resolution using OpenCV bilinear interpolation. The $win_{edge}$ SLIDE is set at 3 sec. Table VIII shows the performance of CloudSeg and VID-WIN at different filtering stages. The CloudSeg 4X and 8X achieve bandwidth savings of 93.76% and 98.43% respectively, whereas VID-WIN filtering with difference values and compression achieves bandw-

| Technique | Bandwidth Saving | Batch Latency (sec) | Avg. Throughput (fps) | Avg. Model Accuracy |
|---|---|---|---|---|
| CloudSeg 4X | 93.76% | 45.87 (batch size =1) | 0.021 | 0.767 |
| CloudSeg 8X | 98.43% | 23.72 (batch size =1) | 0.04 | 0.751 |
| VID-WIN+ No filtering | 97.58% | 3.77 | 18.47 | 0.641 |
| VID-WIN+ filtering (eager) | 97.92% | 3.59 | 17.37 | 0.638 |
| VID-WIN+ filtering(eager+lazy) | 98.71% | 3.16 | 12.76 | 0.631 |
| VID-WIN+ filtering(eager+lazy) diff+compression | 99.12% | 3.18 | 12.41 | 0.631 |

-idth savings of 99.12%. The throughput (0.04 fps) and latency (23.72 sec) of CloudSeg 8X are extremely low as the processing occurs across two models, i.e. CARN to upsample the low-resolution frame and then ResNet101 for classification. The slow processing of the CARN model increases the backpressure queueing, leading to high latency and low throughput. VID-WIN (no filtering) and VID-WIN (filtering + compression) achieve a throughput of 461.7X and 310.2X higher than CloudSeg 8X. The throughput decreases due to filtering as only a fewer number of frames are processed in a given time to save resources.

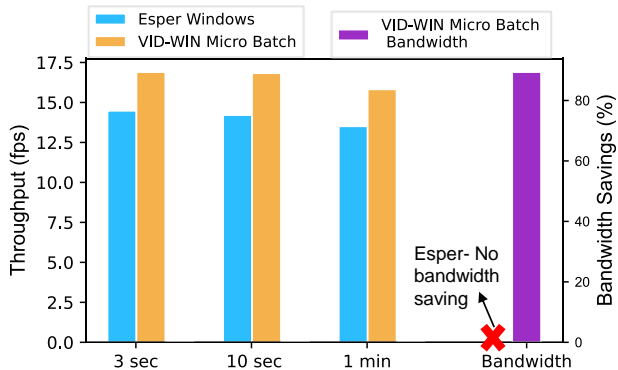

Fig. 24. Performance comparison of VID-WIN with Esper CEP windows on Jackson Hole dataset for throughput and bandwidth savings.

### 5) VID-WIN comparison with Esper CEP engine

Esper is a well-known CEP and stream analytics platform. We have benchmarked the Esper windows with VID-WIN to identify the system performance. The Esper windows are rigid and cannot be decoupled like VID-WIN. Esper runs on the high-end node as in Scenario 2 (Section V-A) where windows receive the frames and pass them to the DNN model. Due to the complexity of retraining the DNN model in Java, we have used the pre-trained ResNet50 model available from the Deeplearning4j zoo library. The pre-trained model comes with a restriction that it only accepts frame resolution of (224,224). So, we have compared only VID-WIN micro-batching performance with a fixed resolution of (224,224). A custom object classification operator is created which is used in Esper EPL query language for the evaluation. Fig. 24 shows the Esper windows performance with VID-WIN. There are no bandwidth savings in the Esper setting as no window is running on an edge device, while VID-WIN micro-batching (with diff+compression) saves 89.4% of the bandwidth. VID-WIN micro-batching achieves a throughput of 16.9, 16.8 and 15.8 fps for a window size of 3 sec, 10 sec and 1 minute. The VID-WIN micro-batch average throughput is 1.2X times higher than Esper, and the performance will increase significantly if resizing and filtering are also considered in the overall experimentation cycle.

| Techniques | Bandwidth Savings (%) | Model Average Accuracy |
|---|---|---|
| Post Processing only ($win_{cloud}$) | 53.67 | 0.796 |
| VID-WIN (Microbatch+resizing+ diff + compression) only [($win_{edge}$)] | 91.4 | 0.676 |
| VID-WIN (Microbatch+resizing+diff+ compression) [($win_{edge}$)] + Post Processing ($win_{cloud}$) | 92.18 | 0.691 |

### 6) VID-WIN post-processing performance

Table IX shows the post-processing performance of VID-WIN. The interesting regions (Q1-car) were continuously extracted using the Faster R-CNN model over 10 minutes on the Sandy Lane dataset which is later sent to the edge node to forward only the selected region. If just ROI post-processing optimization ($win_{cloud}$) is applied, then it can save up to 53.67% of bandwidth as compared to HD streaming with an average model accuracy of ~0.8. The VID-WIN micro-batch resizing using $win_{edge}$ achieves bandwidth savings of up to 91.8% which can further enhance up to 92.18% if post-processing is included in the processing. The inclusion of post-processing in VID-WIN also increases the overall model accuracy from 0.676 to 0.691. This is because the model receives only the region where query-based objects are available, reducing the overall noise which it receives from the non-interested areas.

### C. Limitations

Device enhanced Mobile Edge Computing (MEC) [80, 81] is one of an evolving domain where computations are offloaded to the cluster of mobile devices. These devices share their compute and storage resources and collaborate to provide data-intensive services with low latency. The current VID-WIN design formation is limited to the edge-cloud scenario and cannot be deployed in edge-to-edge MEC settings. This is due to the computationally intensive nature of videos and state management requirements where patterns can occur over long durations. VID-WIN sends optimized micro-batches of video frames to DNN models (such as Faster-RCNN object detector) for further processing. Such DNN based object detection models are resource hungry and have significantly less support where their network architecture can be deployed in distributed MEC settings. The same limitation applies to ad hoc decentralized clouds [82] as VID-WIN requires at least one compute-intensive node to handle the DNN model and large video state depending on the query requirements. Thus VID-WIN approach works in pairs where one window runs on less compute node (edge devices) while the other on a higher compute node (mostly cloud). This way, VID-WIN utilizes the available edge resource budget to improve QoS metrics like throughput, latency, bandwidth usage for state-based video event matching under desired query results. Following the above condition, if windows pairing is maintained, then the approach can be rescaled to multiple nodes depending on the state management and query requirements.

## IX. RELATED WORK

### A. Stream Processing and CEP Optimization

Work from [83], [84] and Drizzle [85] have studied the effect of dynamic batch sizes for adaptive stream processing but were limited to structured data. At the same time, VID-WIN focuses on dynamic batching over unstructured video streams. VideoStorm [45] and Chameleon [44] processes video streams by tuning knobs as per resource quality tradeoffs on large clusters with no focus on edge deployments, windows and state management. Zhao et al. [86] and hSPICE [87] performed state-based load shedding in CEP but limited to structured events while VID-WIN performs state-based filtering over video data in CEP.

### B. Optimization over Windows

Aggregation over windows is one of the research foci where an aggregate operator like SUM, MIN, and machine learning models are applied over an incoming stream. Different aggregation techniques like sharing [17], slicing [6], and multi-query optimization [7] have been proposed for utilizing window state. Bifet et al. [2] proposed ADWIN, content-driven windows where window length changes as per change in data distribution rate. Later Carbone et al. [6] applied the idea of content-driven and aggregate sharing in user-defined windows. All these works consider the incoming data stream as having a fixed data model with a structured payload and are not focused on optimizations related to unstructured content like videos. Yadav et al. have proposed query-aware windows optimizations such as VEKG-TAG [88, 89] and VEKG-EAG [90] for video streams. Their work was focused on the aggregation of graph streams to optimize CEP matching. On the other hand, VID-WIN focuses on low-video data to accelerate overall system performance in the edge-cloud paradigm. The VEKG aggregation technique can be applied at $win_{cloud}$ to further improve system performance.

### C. Edge-based Video Analytics Optimization

FilterForward [49] uses micro classifier, a set of lightweight filters over an edge device to transmit only relevant frames for bandwidth savings. It performs frame-level filtering and does not focus on state-based pattern filtering which is the core of the VID-WIN approach. Reducto [15] performs on camera filtering using cheap vision filters. It saves bandwidth and reduces latency using filtering but do not focus on throughput and state management. Wang et al. [24] proposed different strategies to reduce transmission and save bandwidth but have not considered query awareness, state management, resource allocation. mVideo [7], Vigil [91], AWStream [50] and EdgeEye [92] are edge-based video analytics systems that focus on bandwidth saving while maintaining query accuracy with no focus on state management. MCDNN [25] proposes a scheduling algorithm to select a specialized model trained at different resource requirements. The VID-WIN approach uses standard DNN models to tradeoff resource requirements with query accuracy. NoScope [8], BlazeIt [29], and Focus [67] try to accelerate model inference using specialized DNN models over databases. This work focusses on analyzing content from video streams over windows to increase CEP system performance.

## X. CONCLUSION

In this work, a distributed video event-based adaptive *windowing technique* is presented which is deployed across the *edge and high-end resources* to support the CEP *queries*. This paper presents VID-WIN, a 2-stage allied windowing that runs on edge and cloud. It expands the concept of content-driven widowing from structured data streams to unstructured video data via the input transformation technique. VID-WIN proposes a dynamic micro-batch resizing approach that improves inference time for state-based CEP matching. The approach constitutes a query, state, and resource-aware filtering to amortize edge resource cost with significant bandwidth savings. Extensive experiments on real-world video datasets show the efficacy of VID-WIN based windowing on edge resources and faster execution time with minimal effect on query accuracy. The current work focuses on temporal queries. In future work, we would like to add the spatiotemporal nature of videos in the filtering strategy for a complex event matching over the edge.

## REFERENCES

[1]    S. A. Alvi, B. Afzal, G. A. Shah, L. Atzori, and W. Mahmood, "Internet of multimedia things: Vision and challenges," *Ad Hoc Networks,* vol. 33, pp. 87-111, 2015.

[2]    Y. Lu, A. Chowdhery, and S. Kandula, "Optasia: A relational platform for efficient large-scale video analytics," in *Proceedings of the Seventh ACM Symposium on Cloud Computing*, 2016, pp. 57-70.

[3]    P. Yadav, D. Sarkar, D. Salwala, and E. Curry, "Traffic Prediction Framework for OpenStreetMap Using Deep Learning Based Complex Event Processing and Open Traffic Cameras," in *11th International Conference on Geographic Information Science (GIScience 2021)-Part I*, 2020: Schloss Dagstuhl-Leibniz-Zentrum für Informatik.

[4]    A. R. Elias, N. Golubovic, C. Krintz, and R. Wolski, "Where's the bear?-automating wildlife image processing using iot and edge cloud systems," in *2017 IEEE/ACM Second International Conference on Internet-of-Things Design and Implementation (IoTDI)*, 2017, pp. 247-258: IEEE.

[5]    G. Cugola and A. Margara, "Processing flows of information: From data stream to complex event processing," *ACM Computing Surveys (CSUR),* vol. 44, no. 3, pp. 1-62, 2012.

[6]    G. Cugola and A. Margara, "The complex event processing paradigm," in *Data Management in Pervasive Systems*: Springer, 2015, pp. 113-133.

[7] H. Sun, Y. Yu, K. Sha, and B. Lou, "mVideo: Edge Computing Based Mobile Video Processing Systems," *IEEE Access,* vol. 8, pp. 11615-11623, 2019.

[8] P. Yadav and E. Curry, "VidCEP: Complex Event Processing Framework to Detect Spatiotemporal Patterns in Video Streams," in *2019 IEEE International Conference on Big Data (Big Data)*, 2019, pp. 2513-2522: IEEE.

[9] R. Mayer, B. Koldehofe, and K. Rothermel, "Predictable low-latency event detection with parallel complex event processing," *IEEE Internet of Things Journal,* vol. 2, no. 4, pp. 274-286, 2015.

[10] G. Cugola and A. Margara, "Complex event processing with T-REX," *Journal of Systems and Software,* vol. 85, no. 8, pp. 1709-1728, 2012.

[11] X. Cai, H. Kuang, H. Hu, W. Song, and J. Lü, "Response time aware operator placement for complex event processing in edge computing," in *International Conference on Service-Oriented Computing*, 2018, pp. 264-278: Springer.

[12] J. A. C. Soto, M. Jentsch, D. Preuveneers, and E. Ilie-Zudor, "CEML: Mixing and moving complex event processing and machine learning to the edge of the network for IoT applications," in *Proceedings of the 6th International Conference on the Internet of Things*, 2016, pp. 103-110.

[13] J. Redmon, S. Divvala, R. Girshick, and A. Farhadi, "You only look once: Unified, real-time object detection," in *Proceedings of the IEEE conference on computer vision and pattern recognition*, 2016, pp. 779-788.

[14] Y. Wang, W. Wang, J. Zhang, J. Jiang, and K. Chen, "Bridging the edge-cloud barrier for real-time advanced vision analytics," in *11th USENIX Workshop on Hot Topics in Cloud Computing (HotCloud 19)*, 2019.

[15] Y. Li, A. Padmanabhan, P. Zhao, Y. Wang, G. H. Xu, and R. Netravali, "Reducto: On-Camera Filtering for Resource-Efficient Real-Time Video Analytics," in *Proceedings of the Annual conference of the ACM Special Interest Group on Data Communication on the applications, technologies, architectures, and protocols for computer communication*, 2020, pp. 359-376.

[16] EsperTech. *EsperTech: Event Series Intelligence.* Available: http://www.espertech.com/esper/

[17] A. Arasu and J. Widom, "Resource sharing in continuous sliding-window aggregates," in *Proceedings of the VLDB Endowment*, 2004, pp. 336-347.

[18] J. Li, D. Maier, K. Tufte, V. Papadimos, and P. A. Tucker, "No pane, no gain: efficient evaluation of sliding-window aggregates over data streams," *Acm Sigmod Record,* vol. 34, no. 1, pp. 39-44, 2005.

[19] S. Krishnamurthy, C. Wu, and M. Franklin, "On-the-fly sharing for streamed aggregation," in *Proceedings of the 2006 ACM SIGMOD international conference on Management of data*, 2006, pp. 623-634.

[20] M. Hirzel, S. Schneider, and K. Tangwongsan, "Sliding-window aggregation algorithms: Tutorial," in *Proceedings of the 11th ACM International Conference on Distributed and Event-based Systems*, 2017, pp. 11-14: ACM.

[21] A. Bifet and R. Gavalda, "Learning from time-changing data with adaptive windowing," in *Proceedings of the 2007 SIAM international conference on data mining*, 2007, pp. 443-448: SIAM.

[22] P. Carbone, J. Traub, A. Katsifodimos, S. Haridi, and V. Markl, "Cutty: Aggregate sharing for user-defined windows," in *Proceedings of the 25th ACM International on Conference on Information and Knowledge Management*, 2016, pp. 1201-1210.

[23] Nvidia. *Jetson Nano: Deep Learning Inference Benchmarks.* Available: https://developer.nvidia.com/embedded/jetson-nano-dl-inference-benchmarks

[24] J. Wang *et al.*, "Bandwidth-efficient live video analytics for drones via edge computing," in *2018 IEEE/ACM Symposium on Edge Computing (SEC)*, 2018, pp. 159-173: IEEE.

[25] S. Han, H. Shen, M. Philipose, S. Agarwal, A. Wolman, and A. Krishnamurthy, "Mcdnn: An approximation-based execution framework for deep stream processing under resource constraints," in *Proceedings of the 14th Annual International Conference on Mobile Systems, Applications, and Services*, 2016, pp. 123-136.

[26] C. Pakha, A. Chowdhery, and J. Jiang, "Reinventing video streaming for distributed vision analytics," in *10th USENIX Workshop on Hot Topics in Cloud Computing (HotCloud 18)*, 2018.

[27] P. Yadav, "High-performance complex event processing framework to detect event patterns over video streams," in *Proceedings of the 20th International Middleware Conference Doctoral Symposium*, 2019, pp. 47-50.

[28] S. Chakravarthy, A. Aved, S. Shirvani, M. Annappa, and E. Blasch, "Adapting stream processing framework for video analysis," *ICCS 2015 International Conference On Computational Science, Procedia Computer Science,* vol. 51, pp. 2648-2657, 2015.

[29] D. Kang, P. Bailis, and M. Zaharia, "BlazeIt: optimizing declarative aggregation and limit queries for neural network-based video analytics," *Proceedings of the VLDB Endowment,* vol. 13, no. 4, pp. 533-546, 2019.

[30] P. Garcia Lopez *et al.*, "Edge-centric computing: Vision and challenges," *ACM SIGCOMM Computer Communication Review,* 2015.

[31] C. Savaglio and G. Fortino, "A Simulation-driven Methodology for IoT Data Mining Based on Edge Computing," *ACM Transactions on Internet Technology,* vol. 21, no. 2, pp. 1-22, 2021.

[32] G. Fortino, C. Savaglio, G. Spezzano, and M. Zhou, "Internet of Things as System of Systems: A Review of Methodologies, Frameworks, Platforms, and Tools," *IEEE Transactions on Systems, Man, Cybernetics: Systems,* 2020.

[33] C. Savaglio, P. Gerace, G. Di Fatta, and G. Fortino, "Data mining at the IoT edge," in *2019 28th International Conference on Computer Communication and Networks (ICCCN)*, 2019, pp. 1-6: IEEE.

[34] H. Gupta, A. Vahid Dastjerdi, S. K. Ghosh, and R. Buyya, "iFogSim: A toolkit for modeling and simulation of resource management techniques in the Internet of Things, Edge and Fog computing environments," *Software: Practice and Experience,* vol. 47, no. 9, pp. 1275-1296, 2017.

[35] X. Zeng, S. K. Garg, P. Strazdins, P. P. Jayaraman, D. Georgakopoulos, and R. Ranjan, "IOTSim: A simulator for analysing IoT applications," *Journal of Systems Architecture,* vol. 72, pp. 93-107, 2017.

[36] D. Kang, J. Emmons, F. Abuzaid, P. Bailis, and M. Zaharia, "NoScope: Optimizing Neural Network Queries over Video at Scale," *Proceedings of the VLDB Endowment,* vol. 10, no. 11, 2017.

[37] R. Xu *et al.*, "ApproxNet: Content and Contention Aware Video Analytics System for the Edge," *arXiv preprint arXiv:.02068,* 2019.

[38] P. M. Grulich and F. Nawab, "Collaborative edge and cloud neural networks for real-time video processing," *Proceedings of the VLDB Endowment,* vol. 11, no. 12, pp. 2046-2049, 2018.

[39] K. He, X. Zhang, S. Ren, and J. Sun, "Spatial pyramid pooling in deep convolutional networks for visual recognition," *IEEE transactions on pattern analysis and machine intelligence,* vol. 37, no. 9, pp. 1904-1916, 2015.

[40] Youtube. (2009). *Sandy Lane Corner Video.* Available: https://www.youtube.com/watch?v=e_WBuBqS9h8&t=620s

[41] K. He, X. Zhang, S. Ren, and J. Sun, "Deep residual learning for image recognition," in *Proceedings of the IEEE conference on computer vision and pattern recognition*, 2016, pp. 770-778.

[42] A. G. Howard *et al.*, "Mobilenets: Efficient convolutional neural networks for mobile vision applications," *arXiv preprint arXiv:.04861,* 2017.

[43] S. Ren, K. He, R. Girshick, and J. Sun, "Faster r-cnn: Towards real-time object detection with region proposal networks," in *Advances in neural information processing systems*, 2015, pp. 91-99.

[44] J. Jiang, G. Ananthanarayanan, P. Bodik, S. Sen, and I. Stoica, "Chameleon: scalable adaptation of video analytics," in *Proceedings of the 2018 Conference of the ACM Special Interest Group on Data Communication*, 2018, pp. 253-266.

[45] H. Zhang, G. Ananthanarayanan, P. Bodik, M. Philipose, P. Bahl, and M. J. Freedman, "Live video analytics at scale with approximation and delay-tolerance," in *14th USENIX Symposium on Networked Systems Design and Implementation (NSDI 17)*, 2017, pp. 377-392.

[46] M. R. Anderson, M. Cafarella, G. Ros, and T. F. Wenisch, "Physical representation-based predicate optimization for a visual analytics database," in *2019 IEEE 35th International Conference on Data Engineering (ICDE)*, 2019, pp. 1466-1477: IEEE.

[47] D. Kang, J. Emmons, F. Abuzaid, P. Bailis, and M. Zaharia, "Noscope: optimizing neural network queries over video at scale," *arXiv preprint arXiv:1703.02529,* 2017.

[48] A. H. Jiang *et al.*, "Mainstream: Dynamic stem-sharing for multi-tenant video processing," in *2018 USENIX Annual Technical Conference (USENIX ATC 18)*, 2018, pp. 29-42.

[49] C. Canel *et al.*, "Scaling video analytics on constrained edge nodes," *arXiv preprint arXiv:.13536*, 2019.

[50] B. Zhang, X. Jin, S. Ratnasamy, J. Wawrzynek, and E. A. Lee, "Awstream: Adaptive wide-area streaming analytics," in *Proceedings of the 2018 Conference of the ACM Special Interest Group on Data Communication*, 2018, pp. 236-252.

[51] C. Zhang, Q. Cao, H. Jiang, W. Zhang, J. Li, and J. Yao, "FFS-VA: A fast filtering system for large-scale video analytics," in *Proceedings of the 47th International Conference on Parallel Processing*, 2018, pp. 1-10.

[52] M. Hirzel, R. Soulé, S. Schneider, B. Gedik, and R. Grimm, "A catalog of stream processing optimizations," *ACM Computing Surveys (CSUR),* vol. 46, no. 4, pp. 1-34, 2014.

[53] J. Li, K. Tufte, D. Maier, and V. Papadimos, "AdaptWID: An adaptive, memory-efficient window aggregation implementation," *IEEE Internet Computing,* vol. 12, no. 6, pp. 22-29, 2008.

[54] P. Yadav, D. P. Das, and E. Curry, "Data-Driven Windows to Accelerate Video Stream Content Extraction in Complex Event Processing," in *Proceedings of the 20th International Middleware Conference Demos and Posters*, 2019, pp. 15-16.

[55] P. Pietzuch, J. Ledlie, J. Shneidman, M. Roussopoulos, M. Welsh, and M. Seltzer, "Network-aware operator placement for stream-processing systems," in *22nd International Conference on Data Engineering (ICDE'06)*, 2006, pp. 49-49: IEEE.

[56] H. Li, Y. Wang, H. Wang, and B. Zhou, "Multi-window based ensemble learning for classification of imbalanced streaming data," *World Wide Web,* vol. 20, no. 6, pp. 1507-1525, 2017.

[57] P. Bhatotia, U. A. Acar, F. P. Junqueira, and R. Rodrigues, "Slider: incremental sliding window analytics," in *Proceedings of the 15th international middleware conference*, 2014, pp. 61-72.

[58] A. Arasu, S. Babu, and J. Widom, "The CQL continuous query language: semantic foundations and query execution," *The VLDB Journal,* vol. 15, no. 2, pp. 121-142, 2006.

[59] *Jackson Hole Wyoming USA Town Square Live Cam.* Available: https://www.youtube.com/watch?v=1EiC9bvVGnk

[60] I. Koprinska and S. Carrato, "Temporal video segmentation: A survey," *J Signal processing: Image communication,* vol. 16, no. 5, pp. 477-500, 2001.

[61] D. Marpe, T. Wiegand, and G. J. Sullivan, "The H. 264/MPEG4 advanced video coding standard and its applications," *IEEE communications magazine,* vol. 44, no. 8, pp. 134-143, 2006.

[62] OpenCV. *Histogram Comparison.* Available: https://docs.opencv.org/3.4/d8/dc8/tutorial_histogram_comparison.html

[63] T. Bouwmans, F. Porikli, B. Höferlin, and A. Vacavant, *Background modeling and foreground detection for video surveillance.* CRC press, 2014.

[64] R. Tavenard *et al.*, "Tslearn, A Machine Learning Toolkit for Time Series Data," *Journal of Machine Learning Research,* vol. 21, no. 118, pp. 1-6, 2020.

[65] H. Ueda, T. Miyatake, and S. Yoshizawa, "IMPACT: an interactive natural-motion-picture dedicated multimedia authoring system," in *Proceedings of the SIGCHI conference on Human factors in computing systems*, 1991, pp. 343-350.

[66] J. Hu. *Security Camera Resolution: A Major Factor Contributes to Crisp Images.* Available: https://reolink.com/security-camera-resolution-comparison-examples/#comparison

[67] K. Hsieh *et al.*, "Focus: Querying large video datasets with low latency and low cost," in *13th USENIX Symposium on Operating Systems Design and Implementation (OSDI 18)*, 2018, pp. 269-286.

[68] M. Everingham, L. Van Gool, C. K. Williams, J. Winn, and A. Zisserman, "The pascal visual object classes (voc) challenge," *International journal of computer vision,* vol. 88, no. 2, pp. 303-338, 2010.

[69] A. Slo, S. Bhowmik, A. Flaig, and K. Rothermel, "pSPICE: Partial Match Shedding for Complex Event Processing," in *2019 IEEE International Conference on Big Data (Big Data)*, 2019, pp. 372-382: IEEE.

[70] M. Bucchi, A. Grez, C. Riveros, and M. Ugarte, "Foundations of Complex Event Processing," *Proceedings of the VLDB Endowment,* vol. 11, no. 2, 2017.

[71] D. Zimmer and R. Unland, "On the semantics of complex events in active database management systems," in *Data Engineering, 1999. Proceedings., 15th International Conference on*, 1999, pp. 392-399: IEEE.

[72] E. Wu, Y. Diao, and S. Rizvi, "High-performance complex event processing over streams," in *Proceedings of the 2006 ACM SIGMOD international conference on Management of data*, 2006, pp. 407-418.

[73] A. Slo, S. Bhowmik, and K. Rothermel, "espice: Probabilistic load shedding from input event streams in complex event processing," in *Proceedings of the 20th International Middleware Conference*, 2019, pp. 215-227.

[74] ZeroMQ. *An open-source universal messaging library.* Available: https://zeromq.org/

[75] *SOUTHAMPTON TRAFFIC CAM.* Available: https://www.youtube.com/watch?v=Z9P_2pCgfBA

[76] *City of Auburn Toomer's Corner Webcam 2.* Available: https://www.youtube.com/watch?v=hMYIc5ZPJL4

[77] *EarthCam Live: Times Square in 4K.* Available: https://www.youtube.com/watch?v=eJ7ZkQ5TC08

[78] T. J. Lee, J. Gottschlich, N. Tatbul, E. Metcalf, and S. Zdonik, "Precision and Recall for Range-Based Anomaly Detection," *In Proc. SysML Conference*, 2018.

[79] Y. Li, E. Agustsson, S. Gu, R. Timofte, and L. Van Gool, "Carn: Convolutional anchored regression network for fast and accurate single image super-resolution," in *Proceedings of the European Conference on Computer Vision (ECCV)*, 2018, pp. 0-0.

[80] M. Mehrabi, D. You, V. Latzko, H. Salah, M. Reisslein, and F. H. Fitzek, "Device-enhanced MEC: Multi-access edge computing (MEC) aided by end device computation and caching: A survey," *IEEE Access,* vol. 7, pp. 166079-166108, 2019.

[81] X. Yang, X. Yu, H. Huang, and H. Zhu, "Energy efficiency based joint computation offloading and resource allocation in multi-access MEC systems," *IEEE Access,* vol. 7, pp. 117054-117062, 2019.

[82] A. J. Ferrer, J. M. Marquès, and J. Jorba, "Towards the decentralised cloud: Survey on approaches and challenges for mobile, ad hoc, and edge computing," *ACM Computing Surveys,* vol. 51, no. 6, pp. 1-36, 2019.

[83] T. Das, Y. Zhong, I. Stoica, and S. Shenker, "Adaptive stream processing using dynamic batch sizing," in *Proceedings of the ACM Symposium on Cloud Computing*, 2014, pp. 1-13.

[84] D. Cheng, X. Zhou, Y. Wang, and C. Jiang, "Adaptive scheduling parallel jobs with dynamic batching in spark streaming," *IEEE Transactions on Parallel Distributed Systems,* vol. 29, no. 12, pp. 2672-2685, 2018.

[85] S. Venkataraman *et al.*, "Drizzle: Fast and adaptable stream processing at scale," in *Proceedings of the 26th Symposium on Operating Systems Principles*, 2017, pp. 374-389.

[86] B. Zhao, N. Q. V. Hung, and M. Weidlich, "Load Shedding for Complex Event Processing: Input-based and State-based Techniques," in *2020 IEEE 36th International Conference on Data Engineering (ICDE)*, 2020, pp. 1093-1104: IEEE.

[87] A. Slo, S. Bhowmik, and K. Rothermel, "hSPICE: state-aware event shedding in complex event processing," in *Proceedings of the 14th ACM International Conference on Distributed and Event-based Systems*, 2020, pp. 109-120.

[88] P. Yadav and E. Curry, "VEKG: Video Event Knowledge Graph to Represent Video Streams for Complex Event Pattern Matching," in *2019 First International Conference on Graph Computing (GC)*, 2019, pp. 13-20: IEEE.

[89] P. Yadav, D. Salwala, D. P. Das, and E. Curry, "Knowledge Graph Driven Approach to Represent Video Streams for Spatiotemporal Event Pattern Matching in Complex Event Processing," *International Journal of Semantic Computing,* vol. 14, no. 03, pp. 423-455, 2020.

[90] P. Yadav, D. P. Das, and E. Curry, "State Summarization of Video Streams for Spatiotemporal Query Matching in Complex Event Processing," in *2019 18th IEEE International Conference On Machine Learning And Applications (ICMLA)*, 2019, pp. 81-88: IEEE.

[91] T. Zhang, A. Chowdhery, P. Bahl, K. Jamieson, and S. Banerjee, "The design and implementation of a wireless video surveillance system," in *Proceedings of the 21st Annual International Conference on Mobile Computing and Networking*, 2015, pp. 426-438.

[92] P. Liu, B. Qi, and S. Banerjee, "Edgeeye: An edge service framework for real-time intelligent video analytics," in *Proceedings of the 1st International Workshop on Edge Systems, Analytics and Networking*, 2018, pp. 1-6.

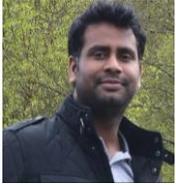 **Piyush Yadav** is a PhD researcher at Insight SFI Research Center for Data Analytics and Lero SFI Software Research Centre at National University of Ireland Galway (NUIG) Ireland. He is in Open Distributed Systems Unit and researching in the field of realtime analytics and pattern detection for Multimedia Data Streams. Before joining NUIG, he was a Researcher at Tata Research Development and Design Centre (TRDDC) which is part of TCS Innovation Lab, India. There he worked in System Robotics Lab and worked in the field of Smart Cities, Urban Modelling and Privacy-Preserving Systems. Before that, he completed his M.Tech. (CSE) with specialization in information security at IIIT Delhi in 2013. His research interests are in Complex Event Processing, Multimedia Analytics, Internet of Things, Machine Learning, Smart Cities, and GIS.

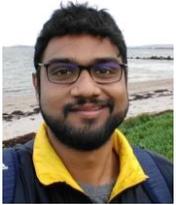 **Dhaval Salwala** is a Research Assistant at the Insight Centre for Data Analytics and Data Science Institute at NUI Galway (NUIG) Ireland since October 2019. He is in the Smart Cities and Sustainable IT group and researching in the field of real-time analytics and pattern detection for Multimedia Data Streams. Before this, he has worked as a Software Professional for over 5 years in the Finance Sector at Tata Consultancy Services (TCS). He has comprehensive experience in designing, developing and deploying E2E architecture for complex business process and data processing pipeline.

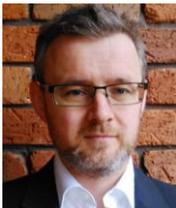 **Edward Curry** is currently a research leader at the Insight SFI Research Centre for Data Analytics and at LERO The Irish Software Research Centre. His research interests are predominantly in open distributed systems, particularly in the areas of incremental data management, approximation, and unstructured event processing, with a special interest in applications for smart environments and data ecosystems. His research work is currently focused on engineering adaptive systems that are a foundation of smart and ubiquitous computing environments. He has authored or coauthored over 150 scientific articles in journals, books, and international conferences. He has presented at numerous events and has given invited talks at Berkeley, Stanford, and MIT. He is a Vice President of the Big Data Value Association—a non-profit industry-led organization with the objective of increasing the competitiveness of European Companies with data-driven innovation. He is a Lecturer in informatics with the National University of Ireland Galway.